\pdfoutput=1

\documentclass[11pt]{article}
\newcounter{mylimit}
\newcommand{\mylim}{\stepcounter{mylimit}(\roman{mylimit})}
\usepackage[preprint]{acl}
\usepackage{longtable}
\usepackage{times}
\usepackage{latexsym}
\usepackage{booktabs}
\usepackage[T1]{fontenc}
\usepackage{amssymb}
\usepackage[utf8]{inputenc}
\usepackage{float} 
\usepackage{microtype}
\usepackage{subfig}
\usepackage{inconsolata}
\usepackage{amsmath}
\usepackage{graphicx}
\usepackage{multirow}
\usepackage{xurl}
\title{Embedding Domain Knowledge for Large Language Models via Reinforcement Learning from Augmented Generation}

\author{
Chaojun Nie$^{1,2}$, Jun Zhou$^{1,2}$\thanks{Corresponding author}, Guanxiang Wang$^{1,2}$ \\
{\bf Shisong Wu$^{3}$, Zichen Wang$^{1,2}$}\\
$^1$Laboratory of Speech and Intelligent Information Processing, \\Institute of Acoustics, Chinese Academy of Sciences, Beijing, China \\
$^2$University of Chinese Academy of Sciences, Beijing, China \\
$^3$China Southern Power Grid Artificial Intelligence Technology Co., Ltd. \\
\{niechaojun, zhoujun, wangguanxiang, wangzichen\}@hccl.ioa.ac.cn, wuss@csg.cn
}

\begin{document}
\maketitle
\begin{abstract}
Large language models (LLMs) often exhibit limited performance on domain-specific tasks due to the natural disproportionate representation of specialized information in their training data and the static nature of these datasets. Knowledge scarcity and temporal lag create knowledge gaps for domain applications. While post-training on domain datasets can embed knowledge into models, existing approaches have some limitations. Continual Pre-Training (CPT) treats all tokens in domain documents with equal importance, failing to prioritize critical knowledge points, while supervised fine-tuning (SFT) with question-answer pairs struggles to develop the coherent knowledge structures necessary for complex reasoning tasks. To address these challenges, we propose Reinforcement Learning from Augmented Generation (RLAG). Our approach iteratively cycles between sampling generations and optimizing the model through calculated rewards, effectively embedding critical and contextually coherent domain knowledge. We select generated outputs with the highest log probabilities as the sampling result, then compute three tailored reward metrics to guide the optimization process. To comprehensively evaluate domain expertise, we assess answer accuracy and the rationality of explanations generated for correctly answered questions. Experimental results across medical, legal, astronomy, and current events datasets demonstrate that our proposed method significantly outperforms baseline approaches. Our code and data are open sourced at \url{https://github.com/ChaojunNie/RLAG}.
\end{abstract}
\section{Introduction}
Large language models (LLMs) have demonstrated exceptional capabilities in capturing and storing factual knowledge across diverse disciplines, attributed to their comprehensive training corpora \citep{roberts2020much,cohen2023crawling,hu2023survey, wang2024knowledge}. However, foundation models trained on broad datasets inherently underrepresent specialized domains relative to their significance in specific applications, creating knowledge gaps in downstream applications. Due to the static nature of training data and the difficulty of accounting for all potential downstream applications during development, LLMs often struggle to answer highly specialized questions \citep{bang2023multitask,ji2023survey,zhang2023sac3}. 
\begin{figure}
    \centering
    \includegraphics[width=1.0\linewidth]{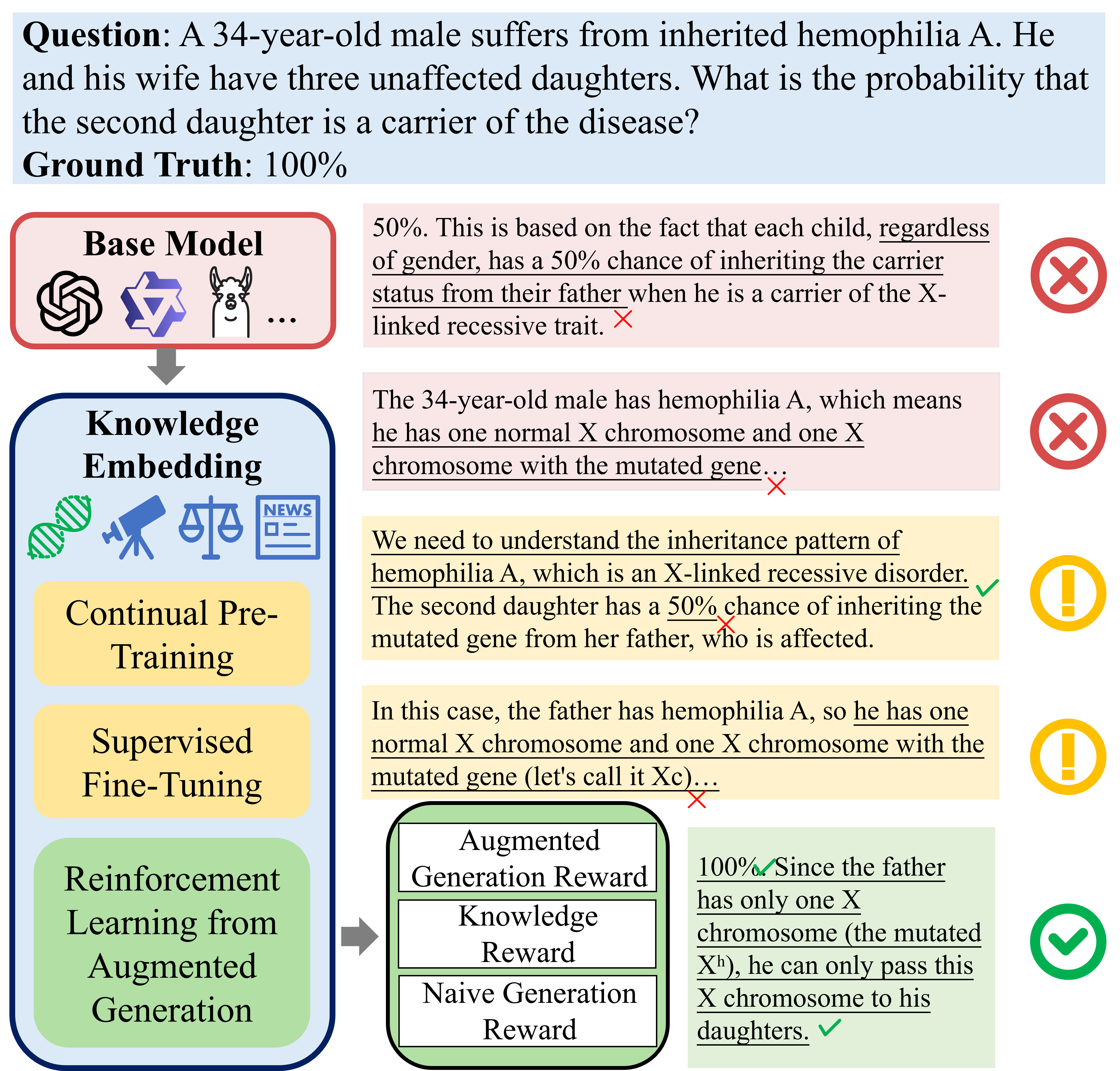}
    \caption{Illustrative example. Base models often struggle with certain task due to limited knowledge. While embedding knowledge into model helps, previous methods may still lead to errors. Our proposed Reinforcement Learning from Augmented Generation (RLAG) incorporates three rewards to optimize models iteratively, improving answer accuracy and explanation rationality.}
    \label{fig1}
\end{figure}
\begin{figure*}
    \centering
    \includegraphics[width=1.0\linewidth]{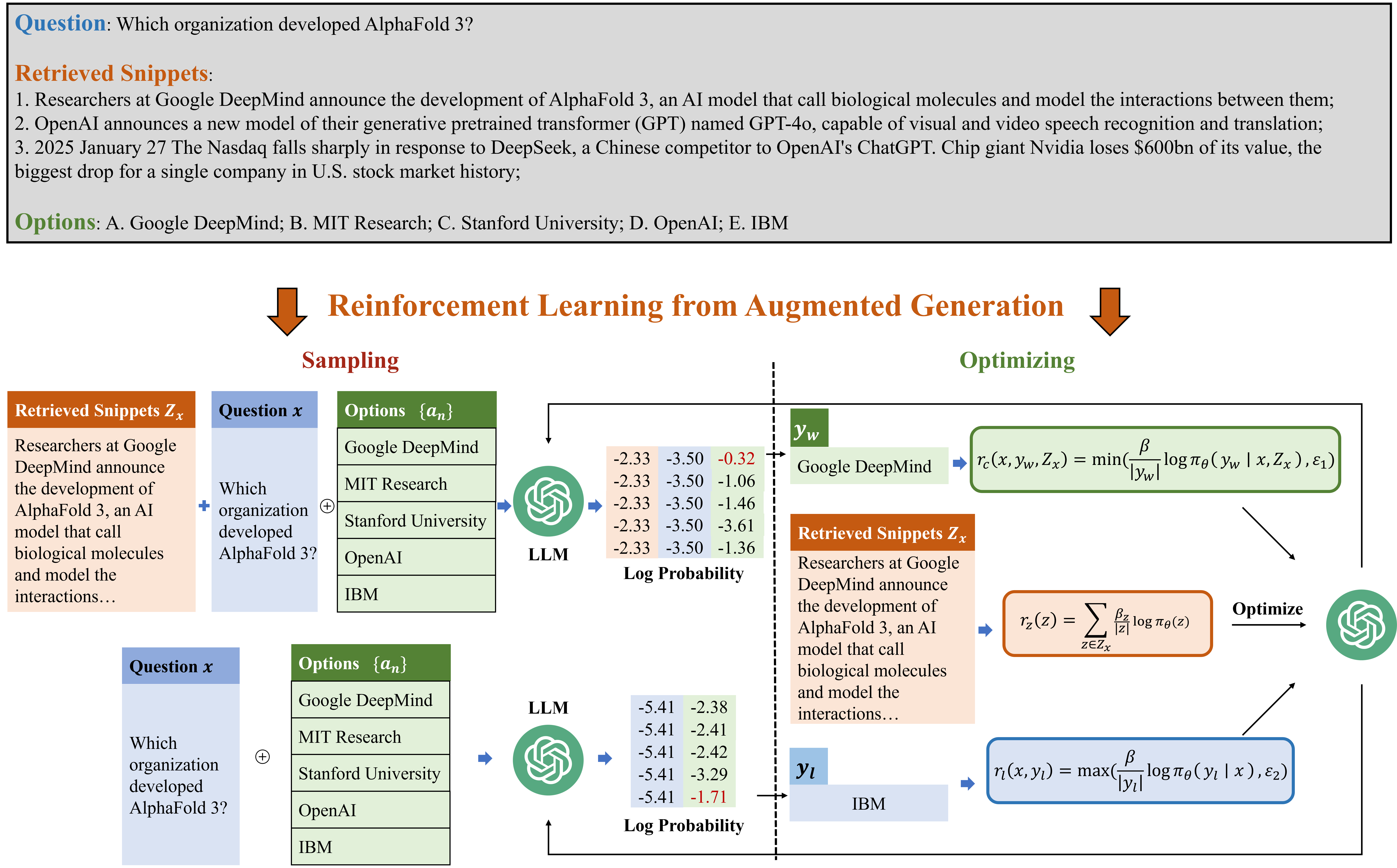}
    \caption{Overview of proposed method: Reinforcement Learning from Augmented Generation (RLAG). Augmented generation $y_w$ (with retrieved snippets $Z_x$) and naive generation $y_l$ (without retrieved snippets) are sampled using Eq~\ref{eq6}. The model is then optimized to increase augmented generation reward $r_c$ and knowledge reward $r_z$ while reducing naive generation reward $r_l$. This process iterates using the updated model for subsequent samples.}
    \label{fig2}
\end{figure*} 

In-context learning (ICL) enhances performance on downstream tasks by providing models with exemplars during inference, enabling adaptation without parameter updates \citep{wang2023labelwordsanchorsinformation,li2023large,highmore2024context}. Retrieval Augmented Generation (RAG) augments model outputs by integrating relevant information from external knowledge bases, improving factual accuracy and reasoning capabilities \citep{guu2020retrieval,lewis2020retrieval,jiang2023active}. Since both ICL and RAG enhance performance through external information at inference time, neither permanently improves the model's intrinsic capabilities for downstream tasks. 

This study focuses on \textit{\textbf{embedding knowledge}} into model weights. Training on downstream datasets embeds domain-specific knowledge directly into model parameters, enabling autonomous reasoning without external support \citep{gururangan2020don,ke2023continual,song2025injecting}.

While Continual Pre-Training (CPT) \citep{ke2023continual} processes entire domain corpora, its effectiveness is limited by the uniform importance assigned to tokens during training \citep{liu2024educating,zhang2024synthetic}. Supervised fine-tuning (SFT) \citep{wei2021finetuned} effectively embeds key information through targeted training; however, models trained exclusively on labeled knowledge pairs often exhibit reduced performance on complex reasoning tasks. 

Inspired by reinforcement learning from human feedback (RLHF) \citep{ouyang2022training, rafailov2023direct}, we introduce \textbf{R}einforcement \textbf{L}earning from \textbf{A}ugmented \textbf{G}eneration (\textbf{RLAG}). In our scenario, generation augmented with relevant literature is preferred over unaugmented generation when addressing downstream questions. The core principle involves \textit{optimizing the model to generate preferred generations independently while continuously improving these generations through iterative refinement}. Notably, our objective extends beyond enabling models to merely reproduce literature-augmented answers (achievable through SFT); we aim for models to thoroughly assimilate knowledge contained within domain literature, thereby maintaining robust knowledge capabilities throughout conversations as shown in Figure~\ref{fig1}. 

As illustrated in Figure~\ref{fig2}, RLAG comprises two principal components: sampling and optimizing. During sampling, we employ a broadcasting operation to concatenate each option with the question, generating two responses differentiated by the presence or absence of retrieved snippets as a prefix. We compute log probabilities for each component through the model's output logits and select the maximum from the option-specific segment as prediction. The optimization phase leverages three predefined reward functions calculated from the sampling results and retrieved snippets to update the model. In the next iteration, we use the updated model for sampling and optimization.

To further isolate LLMs’ abilities to learn new knowledge, we built a dataset covering events post-model training cutoff. Current events dataset is sourced from Wikipedia \citep{wikipedia}. 

We conduct experiments across biomedicine, law, science, and current events. Our experimental results show that the proposed RLAG significantly outperforms prior methods. E.g., in the terms of log-likelihood accuracy, RLAG surpasses prior methods by $14.03\%$ on average on current events dataset.

\section{Preliminaries}
\label{Preferred Ditribution}
In a training iteration, we define output distribution $\pi_{\theta_0}(\cdot\mid x_i,Z_{x_i})$ as the preferred distribution from the LLM (parameters $\theta_0$) conditioned on question $x_i$ and relevant literature $Z_{x_i}$, and $\pi_{\theta_0}(\cdot\mid x_i)$ as the naive distribution conditioned only on question $x_i$.

For a model with parameters $\theta_1$, given problem $x_i$ without relevant literature, its output distribution can match the preferred distribution:
\begin{equation}
    \pi_{\theta_1}(\cdot\mid x_i) \approx \pi_{\theta_0}(\cdot\mid x_i, Z_{x_i}) \succ \pi_{\theta_0}(\cdot\mid x_i)
    \label{eq1}
\end{equation}
The model with parameters $\theta_1$ has internalized downstream knowledge, demonstrating better proficiency than $\theta_0$ \citep{song2025injecting}. When given another downstream problem $x_j$, the distribution $\pi_{\theta_1}(\cdot\mid x_j,Z_{x_j})$ typically outperforms $\pi_{\theta_0}(\cdot\mid x_j,Z_{x_j})$ \citep{ovadia2023fine}:
\begin{equation}
    \pi_{\theta_{1}}(\cdot\mid x_j,Z_{x_j}) \succ \pi_{\theta_0}(\cdot\mid x_j, Z_{x_j})
    \label{eq2}
\end{equation}
Our iterative training objective is to optimize parameters from $\theta_0$ toward $\theta_1$.
\section{Methodology}
\subsection{Sampling}
\label{sample}
During each sampling, the naive generation $y_l$ is sampled from model by concatenating the question $x$ with each option as input, while the augmented generation $y_w$ is sampled from the model with retrieved snippets $Z_x$, question $x$, and each option combined as input, as illustrated in the sampling section of Figure~\ref{fig2}.

For $z'\notin Z_x$, the probability $\pi_\theta(y_w\mid x,z')\approx0$. Thus, $\pi_\theta(y_w\mid x)$ can be approximated as:
\begin{equation}
\label{eq3}
\begin{split}
    \pi_\theta(y_w\mid x) &= \sum_z\pi_\theta(z\mid x)\pi_\theta(y_w\mid x,z) \\
    & \approx \sum_{z\in Z_x}\pi_\theta(z\mid x)\pi_\theta(y_w\mid x,z)
\end{split}
\end{equation}
This decomposition shows that improving $\pi_\theta(y_w\mid x)$ requires increasing either $\pi_\theta(z\mid x)$ or $\pi_\theta(y_w\mid x, z)$. Since $\pi_\theta(y_w\mid x,z)$ is already high and further optimization risks overfitting, we focus on enhancing the posterior probability $\pi_\theta(z\mid x)$.

Directly optimizing $\pi_\theta(z\mid x)$ is computationally challenging. Instead, we enhance the prior probability $\pi_\theta(z)$ to improve $\pi_\theta(z\mid x)$. The relationship between these probabilities is captured by the partial derivative: 
\begin{equation}
    \label{eq4}
    \frac{\partial\pi_\theta(z\mid x)}{\partial\pi_\theta(z)} = \frac{\pi_\theta(x\mid z)\sum_{z'\notin Z_x}\pi_\theta(x\mid z')\pi_\theta(z')}{\pi_\theta(x)^2}
\end{equation}
See Appendix~\ref{eq5_derivation} for a complete derivation.  Since $\pi_\theta(x)>0$ and $\pi_\theta(x\mid z)>0$ for $z\in Z_x$ (as $z$ represents one of the top-$k$ retrieved documents), and making the reasonable and accessible assumption that a sufficiently large document corpus contains at least one relevant snippets $z'\notin Z_x$ with $\pi_\theta(x\mid z')>0$, we can conclude that the derivative is positive:
\begin{equation}
\label{eq5}
    \frac{\partial\pi_\theta(z\mid x)}{\partial\pi_\theta(z)} > 0
\end{equation}
This demonstrates that increasing the prior $\pi_\theta(z)$ effectively enhances the posterior $\pi_\theta(z\mid x)$.

To eliminate prompt template bias, we concatenate each question $x$ with its corresponding options $a_n^l$ and input them into the model, then calculate log probabilities only for the option segment. The prediction is defined as $\mathcal{P}_\theta(x)=c_n$, where:
\begin{equation}
    \label{eq6}
    c_n=\arg \max_l \{\mathcal{P}_\theta(x||a_n^1),\cdots,\mathcal{P}_\theta(x||a_n^L)\}
\end{equation}
and $\mathcal{P}_\theta(x||a_n^l)=\log\pi_\theta(x||a_n^l)$.
\subsection{Reward}
To approximate the target described in Section~\ref{Preferred Ditribution}, we define two reward functions, $r_w$ and $r_l$, which guide the model optimizing. $r_w$ is designed to embed knowledge into model weights and is expressed as:
\begin{equation}
\label{eq7}
\begin{split}
    r&_w(x,y_w, Z_x) \\
    & =\sum_{z\in  Z_x}\frac{\beta_z}{|z|}\log\pi_\theta(z) + \frac{\beta}{|y_w|}\log\pi_\theta(y_w\mid x, Z_x)\\
    & = r_{z}(Z_x) + r_{c}(x,y_w, Z_x)
\end{split}
\end{equation}
where $Z_x$ denotes retrieved snippets relevant to question $x$, and $y_w$ represents the augmented generation. Parameter $\beta_z$ controls the weight of the knowledge reward $r_z$ and $\beta$ adjusts the augmented generation reward $r_c$. Length normalization prevent the model from favoring excessively long outputs. The naive generation reward $r_l$ is defined for naive generation $y_l$ generated without $Z_x$:
\begin{equation}
    r_l(x, y_l) =\frac{\beta}{|y_l|}\log\pi_\theta(y_l\mid x)
    \label{eq8}
\end{equation}
\subsection{Reinforcement Learning from Augmented Generation}
We employ a Bradley-Terry \citep{bradley1952rank} model with target reward margin $\gamma$ \citep{meng2025simpo}. The preference probability is defined as:
\begin{equation}
    P(y_w\succ y_l\mid x)=\sigma(r_w-r_l-\gamma)
    \label{eq9}
\end{equation}
where $\sigma$ denotes the sigmoid function.
\paragraph{Sampling-driven $\beta$ adaption.}
Similar to \text{RLHF}, when sampling yields identical outputs $(y_w=y_l)$, the generation signal becomes invalid, prompting us to set $\beta=0$ to disable generation rewards while retaining the knowledge reward controlled by $\beta_z$. When $y_w\neq y_l$, optimization proceeds with all three rewards activated to optimize model. Full configurations appear in Appendix~\ref{dynamic_beta}.
\paragraph{Clipping strategy.} 
To mitigate overfitting, we introduce a clipping strategy. Probabilities $\pi_\theta(y_w\mid x,Z_x)$ exceeding a threshold $\epsilon_1$ and $\pi_\theta(y_l\mid x)$ falling below a threshold $\epsilon_2$ are clipped.
Substituting $r_w$(Eq~\ref{eq7}),$r_l$(Eq~\ref{eq8}) into Eq~\ref{eq9}. The resulting $\text{RLAG}$ loss function is:
\begin{equation}
\begin{split}
    &\mathcal{L}_{\text{RLAG}} = \\
    &-\mathbb{E}_{(x,y_w,y_l,Z_x) \sim \mathcal{D}} \Bigg[ \log \sigma \Bigg( \sum_{z \in Z_x} \frac{\beta_z}{|z|} \log \pi_\theta(z) \\
    &+ \min \left( \frac{\beta}{|y_w|} \log \pi_\theta(y_w \mid x, Z_x), \epsilon_1 \right) \\
    &- \max \left( \frac{\beta}{|y_l|} \log \pi_\theta(y_l \mid x), \epsilon_2 \right) - \gamma \Bigg) \Bigg].
\end{split}
\label{eq10}
\end{equation}
where $\epsilon_1$ and $\epsilon_2$ are adjustable hyperparameters. The complete derivation appears in the Appendix~\ref{loss_derivation}. Specifically, $\epsilon_1$ caps the maximum probability for augmented generation to avoid overfitting to specific knowledge contexts, while $\epsilon_2$ sets a minimum probability for naive generation to ensure the model does not overly suppress naive generation in the absence of knowledge documents. 
\paragraph{Role of reward components.}
The knowledge reward $r_z$ facilitates the embedding of downstream knowledge into the model by increasing the prior probability of relevant knowledge documents. The augmented generation reward $r_w$ ensures that knowledge embedding aligns with the target parameters, guiding the model toward preferred model. Meanwhile, naive generation reward $r_l$ reduces the likelihood of $y_l$, further reinforcing knowledge integration. Notably, while the generation rewards themselves don't directly embed knowledge into the model, they serve as guides in this optimizing process—architects of direction rather than builders of content.
\begin{table*}
\caption{Results for USMLE \citep{jin2021disease}, BarExamQA \citep{zheng2025reasoning}, and Astronomy \citep{hendrycks2020measuring}. Accuracy quantified by Eq~\ref{eq6}; explanation win rates at temperature 0.3 assessed by GPT-4 Turbo and Grok-3.}
\label{table1}
\resizebox{\textwidth}{!}{
\begin{tabular}{l c c c c c c c c c}
\toprule
\multirow{4}{*}{\textbf{Method}} 
& \multicolumn{9}{c}{\textbf{Llama-3.1-8B-Instruct}} \\
\cmidrule(lr){2-10}
& \multicolumn{3}{c}{USMLE} 
& \multicolumn{3}{c}{BarExamQA} 
& \multicolumn{3}{c}{Astronomy}\\
\cmidrule(lr){2-4}\cmidrule(lr){5-7}\cmidrule(lr){8-10}
& ACC(\%) & GPT-4 Turbo WR(\%) & Grok-3 WR(\%) & ACC(\%) & GPT-4 Turbo WR(\%) & Grok-3 WR(\%) & ACC(\%) & GPT-4 Turbo WR(\%) & Grok-3 WR(\%)\\
\midrule
Base         & 27.6 & 26.2 &  27.2 &  39.3 & \textbf{38.5} & \textbf{39.3} &  46.7 & \textbf{45.4} & 46.7 \\
SFT          & 32.2 & 26.9 &  29.9 & 37.6 & 26.5 & 29.1 &  49.3 & 37.5 & 42.8\\
CPT          & 29.2 & 25.3 &  28.5 & 35.0 & 27.4 & 32.5 &  48.7 & 46.1 & 47.4\\
CPT+SFT     & 33.3 & 25.0 &  30.7 &  36.8 & 26.5 & 23.9 &  48.0 & 38.2 & 42.1\\
\midrule
RLAG          & \textbf{34.8} & \textbf{32.4} &  \textbf{33.9} &  \textbf{41.9} & 35.9 & 38.5 &  \textbf{51.3} & \textbf{45.4} & \textbf{50.0}\\
\toprule
\multirow{4}{*}{\textbf{Method}} 
& \multicolumn{9}{c}{\textbf{Qwen2-7B-Instruct}} \\
\cmidrule(lr){2-10}
& \multicolumn{3}{c}{USMLE} 
& \multicolumn{3}{c}{BarExamQA} 
& \multicolumn{3}{c}{Astronomy}\\
\cmidrule(lr){2-4}\cmidrule(lr){5-7}\cmidrule(lr){8-10}
& ACC(\%) & GPT-4 Turbo WR(\%) & Grok-3 WR(\%) & ACC(\%) & GPT-4 Turbo WR(\%) & Grok-3 WR(\%) & ACC(\%) & GPT-4 Turbo WR(\%) & Grok-3 WR(\%)\\
\midrule
Base         & 25.8 & 21.4 &  24.9 &  34.2 & 32.5 & 32.5 &  50.7 & 50.0 & 50.7 \\
SFT          & 27.7 & 15.0 &  23.2 & 31.6 & 19.7 & 30.8 &  \textbf{53.9} & 50.7 & 50.7\\
CPT          & 26.4 & 21.3 &  25.5 & 35.0 & 32.5 & 35.0 &  48.7 & 46.7 & 46.7\\
CPT+SFT     & 27.0 & 15.9 &  23.3 &  34.2 & 21.4 & 17.9 &  52.0 & 47.4 & 49.3\\
\midrule
RLAG          & \textbf{29.4} & \textbf{23.6} &  \textbf{27.8} &  \textbf{40.2} & \textbf{35.0} & \textbf{38.5} &  53.3 & \textbf{52.0} & \textbf{52.0}\\
\toprule
\multirow{4}{*}{\textbf{Method}} 
& \multicolumn{9}{c}{\textbf{Llama-3.2-3B-Instruct}} \\
\cmidrule(lr){2-10}
& \multicolumn{3}{c}{USMLE} 
& \multicolumn{3}{c}{BarExamQA} 
& \multicolumn{3}{c}{Astronomy}\\
\cmidrule(lr){2-4}\cmidrule(lr){5-7}\cmidrule(lr){8-10}
& ACC(\%) & GPT-4 Turbo WR(\%) & Grok-3 WR(\%) & ACC(\%) & GPT-4 Turbo WR(\%) & Grok-3 WR(\%) & ACC(\%) & GPT-4 Turbo WR(\%) & Grok-3 WR(\%)\\
\midrule
Base         & 26.2 & 22.8 &  24.8 &  34.2 & \textbf{25.6} & 29.1 &  49.3 & 44.1 & 47.4 \\
SFT          & \textbf{30.2} & 25.6 &  27.9 & 33.3 & 21.4 & 24.8 &  50.0 & 46.1 & 48.0\\
CPT          & 27.4 & 22.8 &  25.5 & 28.2 & 16.2 & 22.2 &  47.4 & 40.1 & 44.7\\
CPT+SFT     & 29.3 & 22.3 &  27.2 &  29.9 & 19.7 & 17.1 &  46.7 & 40.8 & 42.1\\
\midrule
RLAG          & 29.7 & \textbf{25.9} &  \textbf{28.1} &  \textbf{36.8} & \textbf{25.6} & \textbf{33.3} &  \textbf{52.0} & \textbf{46.7} & \textbf{51.3}\\
\bottomrule
\end{tabular}%
}
\small
\centering
\caption{Results for the Current Events in terms of log-likelihood accuracy (Eq~\ref{eq6})}
\label{table2}
\begin{tabular}{ l l c c c c c c}
\toprule
\textbf{Task} & \textbf{Model} & \textbf{Base} & \textbf{SFT} & \textbf{CPT} & \textbf{CPT+SFT} & \textbf{RLAG}\\
\midrule
\multirow{3}{*}{Current Events}     & Qwen2-7B-Instruct & 25.3 & 32.1 & 27.2 & 34.6 & \textbf{48.8} \\
                                    & Llama-3.1-8B-Instruct & 30.2 & 34.0 & 29.6 & 35.8 & \textbf{54.9} \\
                                    & Llama-3.2-3B-Instruct & 23.5 & 25.9 & 22.8 & 27.2 & \textbf{37.0} \\
\bottomrule
\end{tabular}
\end{table*}
\section{Knowledge Base Creation}
\subsection{Task Selection and Statistics of Data} 
Experiments were conducted across four distinct downstream tasks. 

\textbf{Biomedicine}: The USMLE task from MedQA \citep{jin2021disease}, drawn from U.S. National Medical Licensing Examinations, represents a high-difficulty challenge in medical reasoning. USMLE comprises $10,178$ training instances, $1,272$ validation instances, $1,273$ testing instances, and $18$ biomedicine books. 

\textbf{Law}: The BarExamQA \citep{zheng2025reasoning} task comprises legal questions from practical bar exams. BarExamQA incorporates $954$ training instances, $124$ validation instances, $117$ testing instances, and legal documents. 

\textbf{Astronomy}: Astronomy task from the MMLU \citep{hendrycks2020measuring} benchmark, with training data generated using GPT-4 Turbo \citep{hurst2024gpt} and DeepSeek-R1 \citep{guo2025deepseek}. This tested the model’s scientific knowledge. The astronomy task contains $2,000$ training instances, $134$ validation instances, and $152$ testing instances. 

\textbf{Current Events}: We developed a dataset encompassing post-training temporal phenomena, consisting of $1,300$ training instances, $169$ validation instances, and $162$ testing instances.

The original developers released these research-focused datasets, which have been extensively cited in academic literature. We strictly comply with each dataset's usage terms, ensuring their application remains limited to scholarly research.
\subsection{Knowledge Base Creation} 
\label{section4.2}

\textbf{USMLE}: For the USMLE task, we curated a knowledge base from $18$ biomedical textbooks provided by the MedQA \citep{jin2021disease} through systematic text cleaning and structural normalization. The USMLE knowledge base (KB) has $17.3$M tokens. All token counts use LlamaTokenizer. 

\textbf{BarExamQA}: For the BarExamQA \citep{zheng2025reasoning} task, we utilized gold passages provided with each sample as reference documents. The BarExamQA KB has $93.1$M tokens.

\textbf{Astronomy}: For the MMLU astronomy task \citep{hendrycks2020measuring}, we followed a structured process: DeepSeek-R1 \citep{guo2025deepseek} extracted keywords from astronomy questions. Then we collected text by searching keywords with the Wikipedia API \footnote{\url{https://www.mediawiki.org/wiki/API:Main_page}} and generated samples using Deepseek-R1. The Astronomy KB has $3.1$M tokens. Following previous work \citep{guo2024deepseek}, we removed questions from the training set that had 3-gram overlaps with the test set to prevent test set contamination. Curation was performed via the Claude-3.7-Sonnet API\footnote{\url{https://www.anthropic.com/claude/sonnet}} and manual review eliminated ambiguous/incorrect questions. 

\textbf{Current events}: For the current events task, we collected events after the model training data cutoff date from Wikipedia \citep{wikipedia}, including: 2024-2025 U.S. events, 2025 German federal election, and 2024 Summer Olympics. The text was segmented and cleaned using spaCy \citep{honnibal2020spacy}. The Current events KB has $51.5$K tokens. GPT-4 Turbo \citep{hurst2024gpt} generated questions for each five-line segment. Equal samples were generated per topic, with test sets uniformly sampled across all events. Recognizing RLAG's potential privacy risks from personal information in training data, we manually screened the dataset to eliminate ethical concerns. This dataset is for academic research use only.
\begin{figure*}[htbp]
    \centering
    \includegraphics[width=1.0\linewidth]{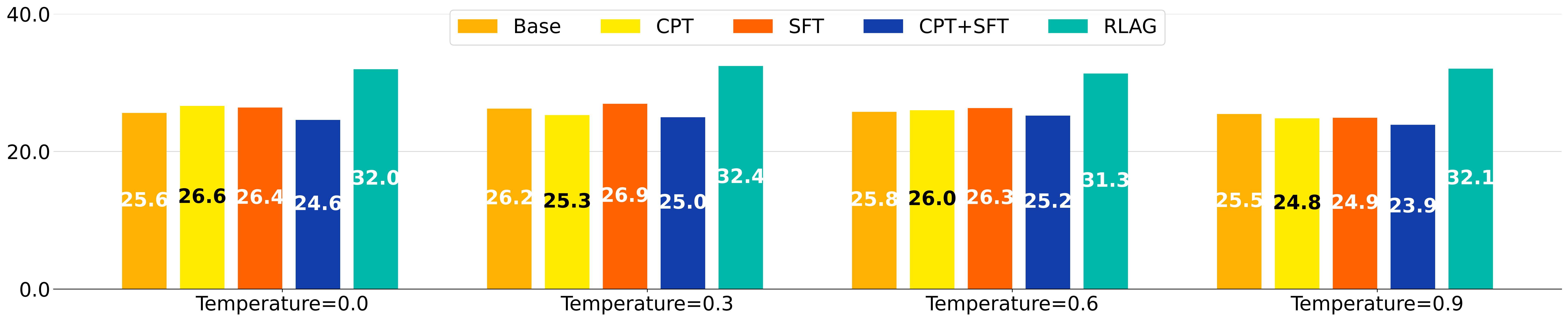}
    \caption{Evaluate explanation quality of questions correctly answered by Llama-3.1-8B-Instruct across temperatures on USMLE dataset, which is conducted by GPT-4 Turbo.}
    \label{fig3}
\end{figure*}
\begin{figure}[htbp]
    \centering
    \includegraphics[width=1.0\linewidth]{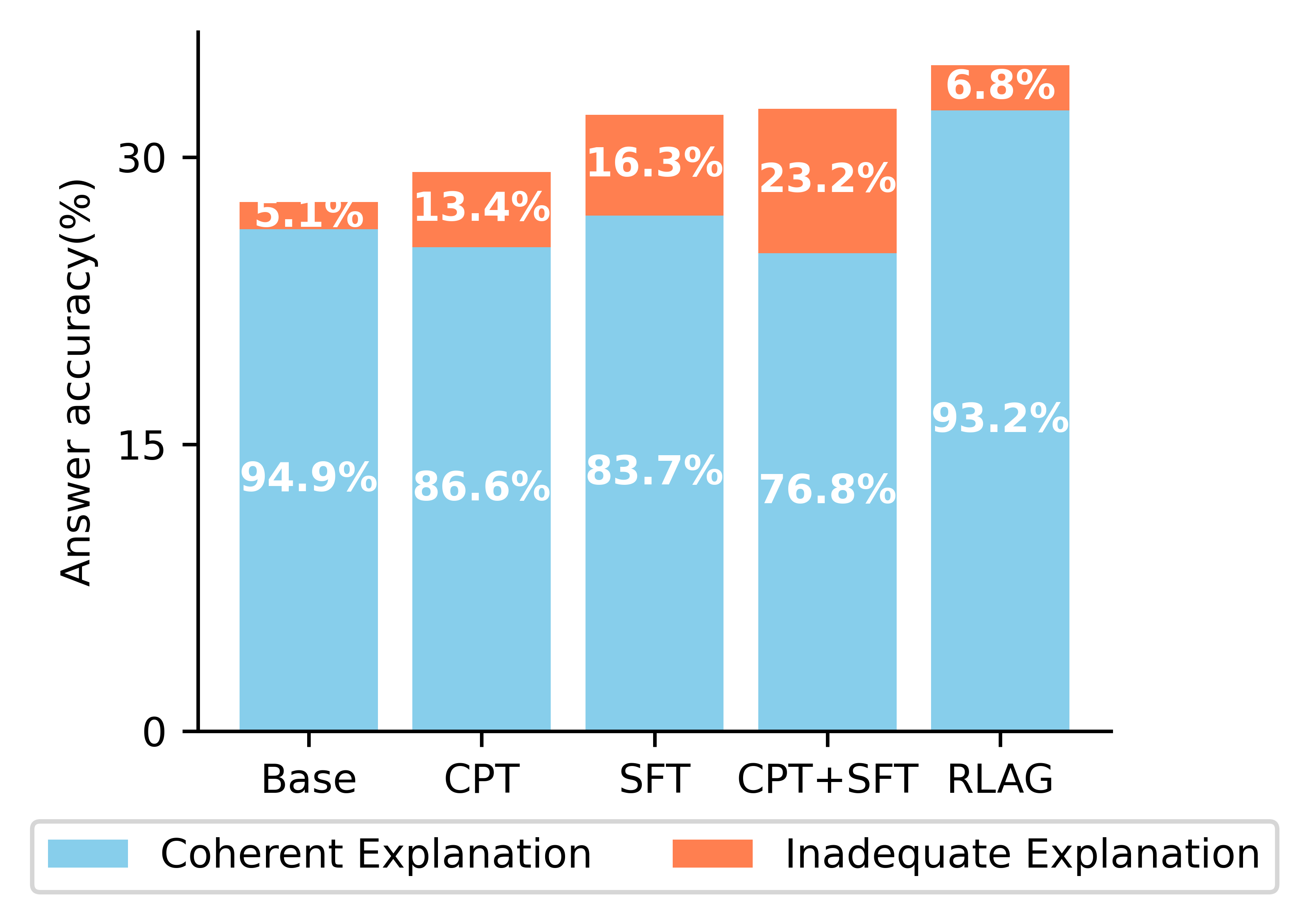}
    \caption{Performance comparison between RLAG and baseline approaches on the USMLE dataset with temperature set to 0.3. Results show answer accuracy and explanation rationality, with explanations evaluated by GPT-4 Turbo.}
    \label{fig4}
\end{figure}
\section{Experiments and Results}
\subsection{Experimental Setup}
\paragraph{Models and training settings.}
Knowledge embedding experiments used two model families: Qwen2 \citep{yang2024qwen2technicalreport} and Llama3 \citep{grattafiori2024llama}. We selected both large and small variants: Qwen2-7B-Instruct, Llama-3.1-8B-Instruct, and Llama-3.2-3B-Instruct to analyze knowledge embedding effects across different parameter scales. We used instruction-tuned models off-the-shelf, as these are commonly deployed in practice, making the embedding of downstream knowledge into these models practically significant. $\textit{NV-Embed-v2}$ \citep{lee2024nv} was selected as the embedding model, and used FAISS \citep{johnson2019billion} as its vector-store. We report the best performance obtained via a grid search of hyperparameters, while ablation studies and evaluation of explanations were conducted with single experimental runs. \\Tokenizers configured with padding token to the end-of-sequence token and assigned Qwen2Tokenizer's beginning-of-sequence token to $<|\text{im\_start}|>$\footnote{\url{https://huggingface.co/docs/transformers/main/en/chat_templating}}. Details of training configurations and retrieval methods are provided in Appendix~\ref{appendix_A.1} and Appendix~\ref{retri method}, respectively. Retrieval ablation experiments in Appendix~\ref{retrieval ablation} show RLAG's robust performance across different retrieval methods.
\paragraph{Baselines.} 
The $\text{SFT}$ loss function is defined as:
\begin{equation}
    \mathcal{L}_{\text{SFT}} = -\sum_{i=1}^{B}\frac{1}{|y_i|}\sum_{j=1}^{|y_i|}\pi_\theta(y_{i,j}\mid x,y_{i,<j})
    \label{eq11}
\end{equation}
where $B$ is the batch size, $y_i$ is the answer sequence, and $y_{i,j}$ is its $j$-th token. We apply length normalization to prevent bias toward longer outputs.

The $\text{CPT}$ loss function is:
\begin{equation}
    \mathcal{L}_{\text{CPT}} = -\sum_{i=1}^{B}\frac{1}{|z_i|}\sum_{j=1}^{|z_i|}\pi_\theta(z_{i,j}\mid z_{i,<j})
    \label{eq12}
\end{equation}
where $z_i$ represents a knowledge document chunk, and $z_{i,j}$ is its $j$-th token. \\To enhance knowledge embedding effectiveness, we also explored a pipeline combining $\text{CPT}$ on knowledge documents followed by $\text{SFT}$.
\subsection{Evaluation Method}
We employed a two-stage sequential evaluation: answer accuracy followed by explanation assessment for correctly answered questions.
\paragraph{Log-likelihood accuracy.}
We ensured prompt-independent results by connecting each option to the question, calculating generation probabilities, and selecting the highest-probability option as the prediction (Eq~\ref{eq6}).
\paragraph{Explanation win rates.} For correctly answered questions, we evaluated knowledge embedding by prompting models to explain their answers. Explanations were assessed for logical clarity and factual accuracy using GPT-4 Turbo \citep{hurst2024gpt} and Grok-3 \citep{xai2024grok3}, with win rates calculated as percentages. Complete evaluation templates appear in Appendix~\ref{eva_tem}.
\begin{table*}[htbp]
\centering
\caption{Ablation study on Llama-3.1-8B-Instruct. We ablate four keys of RLAG:(1) No Clipping in Eq~\ref{eq10} ($i.e., w/o$ Clip), (2) Fix $\beta,\beta_z$ in Eq~\ref{eq10} ($i.e., \text{Fixed }\beta,\beta_z$), (3) Set $\gamma=0$ in Eq~\ref{eq10} ($i.e.,\text{w/o }\gamma$), (4) Replace sample $y_w$ with standard answer in Eq~\ref{eq10} ($i.e.$ Std. Ans. as $y_w$)}
\label{table3}
\resizebox{\textwidth}{!}{
\begin{tabular}{l c c c c c c c c c}
\toprule
\multirow{4}{*}{\textbf{Method}} 
& \multicolumn{9}{c}{\textbf{Llama-3.1-8B-Instruct}} \\
\cmidrule(lr){2-10}
& \multicolumn{3}{c}{USMLE} 
& \multicolumn{3}{c}{BarExamQA} 
& \multicolumn{3}{c}{Astronomy}\\
\cmidrule(lr){2-4}\cmidrule(lr){5-7}\cmidrule(lr){8-10}
& ACC(\%) & GPT-4 Turbo WR(\%) & Grok-3 WR(\%) & ACC(\%) & GPT-4 Turbo WR(\%) & Grok-3 WR(\%) & ACC(\%) & GPT-4 Turbo WR(\%) & Grok-3 WR(\%)\\
\midrule
CPT+SFT     & 33.3 & 25.0 &  30.7 &  36.8 & 26.5 & 23.9 &  48.0 & 38.2 & 42.1\\
\midrule
RLAG          & 34.8 & 32.4 &  33.9 &  41.9 & 35.9 & 38.5 &  51.3 & 45.4 & 50.0\\
\midrule
w/o Clip & 30.5 & 24.1 & 24.6 & 35.0 & 27.4 & 29.1 & 52.0 & 48.0 & 48.7 \\
Fixed $\beta,\beta_z$ & 32.9 & 29.5 &  30.5 &  32.5 & 22.2 & 29.1 &  48.7 & 46.1 & 46.7 \\
w/o $\gamma$ & 32.1 & 29.1 &  29.6 &  36.8 & 28.2 & 29.1 &  48.7 & 46.1 & 48.0 \\
Std. Ans. as $y_w$ & 31.0 & 4.24 &  5.34 &  35.0 & 29.1 & 31.6 &  49.3 & 46.7 & 49.3 \\
\bottomrule
\end{tabular}
} 
\end{table*}
\begin{figure*}[htbp]
    \centering
    \subfloat[Effect of clipping on $r_l$]{
        \includegraphics[width=0.32\textwidth]{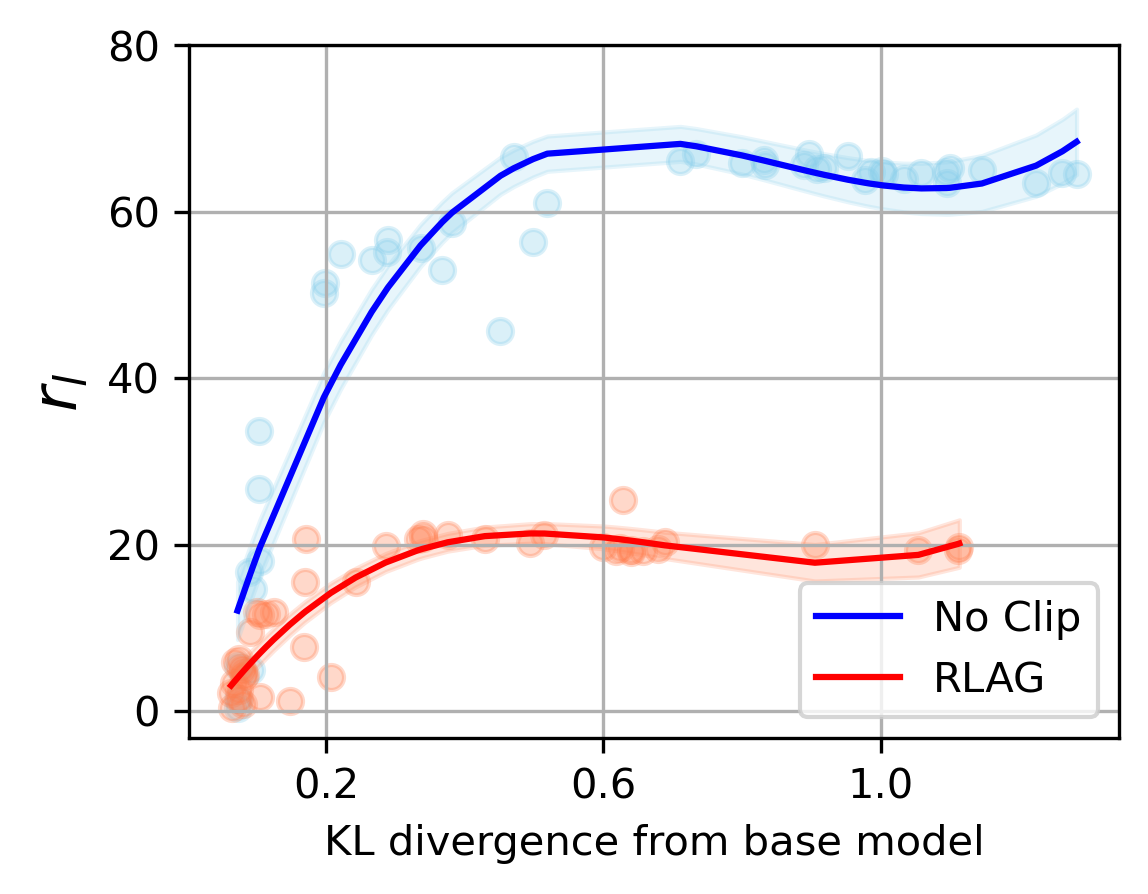}
        \label{fig5a}
    }
    \subfloat[Effect of clipping on $r_w$]{
        \includegraphics[width=0.32\textwidth]{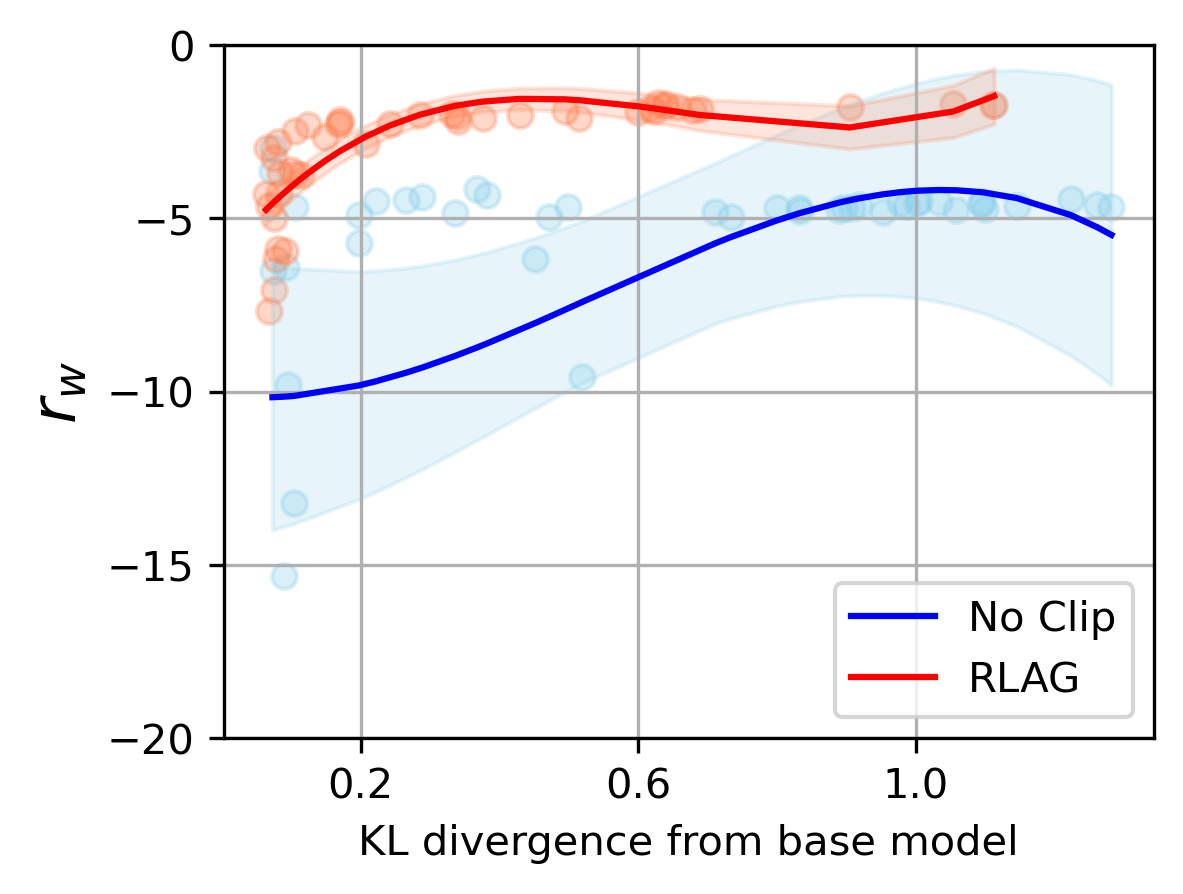}
        \label{fig5b}
    }
    \subfloat[Effect of clipping on validation set]{
        \includegraphics[width=0.32\textwidth]{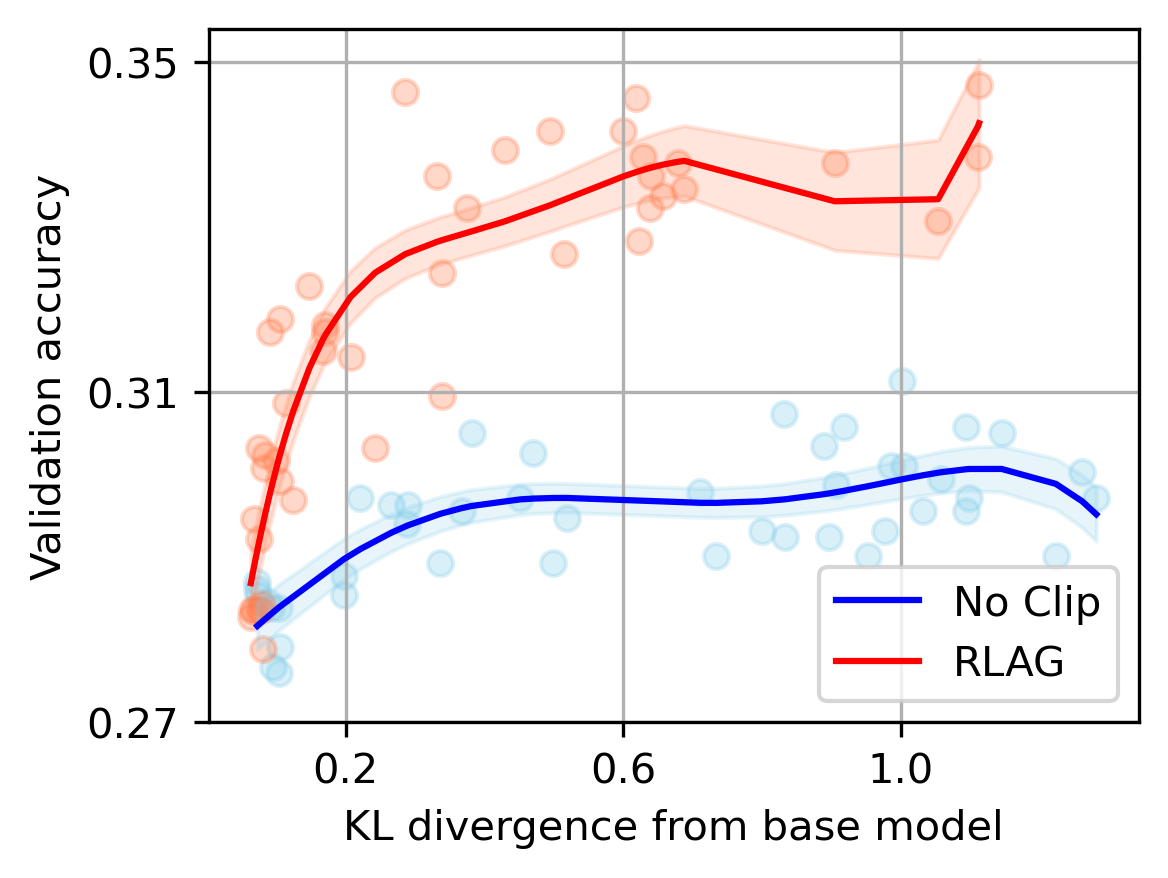}
        \label{fig5c}
    }
    \caption{Ablation study on reward clipping effects: constraining naive reward $r_l$ inflation (a) while steadily increasing $r_w$ (b) and preserving accuracy (c), demonstrating effective reward control.}
    \label{fig5}
\end{figure*}
\subsection{Main Results}
\label{results}
\paragraph{Downstream tasks results.}
Table~\ref{table1} demonstrates RLAG's superior performance across tasks. On USMLE \citep{jin2021disease}, RLAG achieves the highest overall answer accuracy and surpasses all baselines in explanation win rate by $2.2-5.5$ points. For BarExamQA \citep{zheng2025reasoning}, RLAG outperforms the best baseline by $3.5-5.2$ points in accuracy while maintaining superior explanation rationality. This legal reasoning task reveals the limitations of baseline methods: SFT merely learns question-answer mappings without robust reasoning, while CPT suffers from catastrophic forgetting as vast legal documents. Even on Astronomy \citep{hendrycks2020measuring}, where injected knowledge is primarily factual and benefits SFT, RLAG still outperforms all baselines, whereas CPT on the Astronomy knowledge base degrades model performance.
\paragraph{Explanation win rates across temperature.}
As shown in Figure~\ref{fig3}, RLAG outperforms all baselines by $5.0-7.2$ points in explanation win rate across temperatures. While baseline training improves answer accuracy, it compromises explanation rationality, with unexplained portions rising from $5.1\%$ to $13.4-23.2\%$ ($>100\%$ relative increase). RLAG enhances accuracy while preserving explanation quality, with unexplained portions increasing marginally from $5.1\%$ to $6.8\%$ (Figure~\ref{fig4}). This demonstrates that RLAG embeds domain knowledge comprehensively into the model, ensuring logical coherence without requiring manual annotation. We conducted additional small-scale human evaluation to further substantiate RLAG's effectiveness. The experimental protocol and results are provided in Appendix~\ref{small-scale-human-eva}.
\paragraph{Current events results.}
Table~\ref{table2} presents results on current events. Although CPT+SFT pipeline can effectively improve the performance of the model, RLAG demonstrates significant gains of $9.8-19.1$ points over optimal baselines. Larger 7B-8B models show more substantial improvements ($14.2$ and $19.1$ points respectively), while the 3B model improves by $9.8$ points. As this task focuses on factual questions, explanation rationality was not evaluated.
\subsection{Ablation Studies}
\label{ablation}
Four components were evaluated in RLAG via ablation studies with Llama-3.1-8B-Instruct: (1) Reward Clipping (w/o Clip), (2) Dynamic $\beta, \beta_z$ (Fixed $\beta,\beta_z$), (3) reward margin $\gamma$ (w/o $\gamma$), and (4) directly using the standard answer as the augmented generation in Eq~\ref{eq10} (Std. Ans. as $y_w$).

Table~\ref{table3} shows all components are critical, with reward clipping having the strongest impact. Removing reward clipping significantly affects reasoning tasks, reducing answer accuracy by $4.3\%$ on USMLE and $7.9\%$ on BarExamQA, with explanation rationality decreasing by $9\%$ for both tasks. However, it minimally impacts factual knowledge tasks like Astronomy. Fixed $\beta,\beta_z$ and removing reward margin $\gamma$ also decrease performance. Using standard answers as augmented generation (Eq~\ref{eq10}) dramatically reduces performance, causing serious hallucinations—USMLE explanation rationality drops by over $28$ points. This indicates models may learn correct answers but fail to develop robust reasoning when answers are directly provided rather than autonomously generated.
\paragraph{The key role of reward clipping.}
Reward clipping is essential in our method. Figure~\ref{fig5a} shows unconstrained naive generation reward $r_l$ rapidly increases as model divergence occurs, while Figure~\ref{fig5b} indicates minimal growth in $r_w$, yielding negligible validation accuracy improvements (Figure~\ref{fig5c}). Conversely, RLAG with reward clipping effectively constrains $r_l$ while maintaining superior $r_w$ compared to the w/o clipping. This results in consistently higher validation accuracy, highlighting reward clipping's critical contribution to model performance.
\paragraph{Using standard answer weakens RLAG.}
Direct substitution of standard answers for augmented generation significantly degrades model performance and induces hallucinations (Table~\ref{table3}), particularly in reasoning-intensive domains like USMLE. Our case study (Appendix~\ref{case_study}) demonstrates that this approach causes the model to contradict previously answered questions and question the validity of given options. The effectiveness of knowledge embedding strategies ultimately depends on task complexity and reasoning requirements.
\begin{table}[htbp]
    \centering
\caption{Computational Budget in terms of GPU hours}
\label{table4}
\resizebox{\linewidth}{!}{
\begin{tabular}{ l l c c c c c}
\toprule
\textbf{Task} & \textbf{Model} & \textbf{SFT} & \textbf{CPT} & \textbf{CPT+SFT} & \textbf{RLAG}\\
\midrule
\multirow{3}{*}{USMLE}     & Qwen2-7B-Instruct  & 4 & 4 & 8 & 32 \\
                                    & Llama-3.1-8B-Instruct  & 6 & 4 & 10 & 34 \\
                                    & Llama-3.2-3B-Instruct  & 3 & 2 & 5 & 18 \\
\midrule
\multirow{3}{*}{BarExamQA}     & Qwen2-7B-Instruct & 1 & 22 & 23 & 8 \\
                                    & Llama-3.1-8B-Instruct & 1 & 27 & 28 & 9 \\
                                    & Llama-3.2-3B-Instruct & 1 & 12 & 13 & 6 \\
\midrule
\multirow{3}{*}{Astronomy}     & Qwen2-7B-Instruct & 1 & 1 & 2 & 12 \\
                                    & Llama-3.1-8B-Instruct & 1 & 1 & 2 & 10 \\
                                    & Llama-3.2-3B-Instruct & 0.3 & 0.3 & 0.6 & 8 \\
\midrule
\multirow{3}{*}{Current Events}     & Qwen2-7B-Instruct & 1 & 0.3 & 1.3 & 8 \\
                                    & Llama-3.1-8B-Instruct & 0.8 & 0.3 & 1 & 8 \\
                                    & Llama-3.2-3B-Instruct & 0.3 & 0.3 & 0.6 & 4 \\
\bottomrule
\end{tabular}}
\end{table}
\subsection{Computational Budget}
All experiments ran on a server with four NVIDIA A800 GPUs (80GB each). As shown in Table~\ref{table4}, RLAG training requires approximately one order of magnitude more GPU hours than the baseline due to online sampling and optimization processes. Future implementation updates could focus on incorporating efficient sampling frameworks such as vLLM \citep{kwon2023efficient}, which would reduce sampling time and thus decrease the overall training time. Despite the increased computational cost, the significant performance improvements justify this additional investment.
\section{Related Work}
\paragraph{Knowledge injection.}
In order to enhance LLMs' capabilities in downstream tasks, knowledge injection is considered a promising research direction \citep{chen2022knowprompt,ye2023qilin}. Knowledge injection for LLMs can occur during pre-training, fine-tuning, or inference stages. Methods include: (1) RAG, which retrieves text \citep{guu2020retrieval,lewis2020retrieval} or graph-structured \citep{wang2023dynamic,zhang2023learning,li2024framework} information during reasoning; (2) Modular adapters, which incorporate domain knowledge through lightweight additional parameters \citep{zhang2023plug,lo2024m2mkd}; (3) Prompt optimization techniques that leverage internal knowledge \citep{wei2022chain}; and (4) Direct weight embedding through CPT \citep{ke2023continual} or SFT, which enhances domain expertise and stability \citep{gururangan2020don,song2025injecting}. Recent advances focus on optimizing knowledge structures \citep{zhang2024synthetic}, implementing gating mechanisms \citep{peinelt2021gibert}, and developing structure-aware training strategies \citep{liu2024educating} .
\paragraph{Reinforcement learning from human feedback (RLHF).}
RLHF technology enhances LLMs' performance using reinforcement learning with preference data \citep{ouyang2022training,stiennon2020learning}. The approach trains a reward model on preference data, then uses PPO to optimize the policy model, significantly improving generation quality \citep{shao2024deepseekmath, guo2025deepseek}. DPO \citep{radford2019language} reparameterizes reward model and directly using preference data to optimize the policy model. RLHF does not focus on embedding knowledge into the model, but improves the output by aligning with humans. 
\section{Conclusion and Future Work}
In this work, we propose RLAG for knowledge embedding. Compared with traditional knowledge embedding methods, RLAG can solve knowledge-intensive tasks that require reasoning. The core idea of RLAG is to enable the model to independently generate augmented generation and optimize these generation through a reward-based approach. The training process is implemented iteratively by sampling and optimization. Experiments show that RLAG outperforms baseline methods. In future work, we aim to dynamically embed knowledge into LLMs, rather than performing offline training.
\section*{Limitations}
RLAG, while showing promising results in embedding knowledge into LLMs, has several limitations. \mylim  Although RLAG eliminates the need for manual annotation during training, it requires knowledge documents relevant to each question. These document fragments can be collected by searching knowledge bases through retrieval systems. However, the quality of these retrieved fragments depends on retriever performance and knowledge base structure, potentially affecting overall system effectiveness. \mylim The training process of RLAG encompasses two phases: sampling and optimization. While we have demonstrated the sampling process to be effective, it may require more computational time than training directly on existing datasets. \mylim Both sampling and training processes within RLAG require access to token probabilities, making our approach unsuitable for closed-source models that do not provide such access. \mylim Due to hardware constraints, our research primarily focuses on language models with 3B, 7B, and 8B parameters and does not extend to larger-scale models that might yield different performance characteristics. \mylim The datasets for law, astronomy, and current events datasets are relatively smaller compared to the medical dataset, which may affect the generalizability of individual dataset results. However, reporting results across multiple datasets collectively enhances the overall credibility of our findings. \mylim This study employs two powerful commercial large language models—GPT-4 Turbo and Grok-3—to evaluate explanation win rates. Although the results demonstrate reasonable reproducibility, the closed-source nature of these models may introduce variability in evaluation outcomes.
\section*{Ethical Considerations}

\paragraph{Data Collection and Privacy}
Our training datasets pose minimal privacy risks: current events data was exclusively sourced from public Wikipedia with manual screening to exclude personal information, while other datasets are established benchmarks from prior research. RLAG's knowledge embedding capabilities could theoretically raise privacy concerns, but our data sources and screening protocols mitigate these risks.

\paragraph{Human Evaluation Protocol}
We conducted small-scale human evaluation with anonymous participants recruited transparently via social media. Participants were volunteers residing in Beijing and Shenzhen, China. All participants provided informed consent after being fully briefed on task requirements, the approximate 3-minute time commitment per annotation, and how their anonymized responses would be used for research evaluation purposes. Compensation was set at 5 CNY per annotation, ensuring fair payment above minimum wage standards. To protect participant privacy, we implemented strict anonymization protocols using random IDs and secure data storage with restricted access. Participation remained entirely voluntary throughout the study, with participants explicitly informed of their right to withdraw at any time without penalty. The evaluation tasks focused exclusively on benign text quality assessment, ensuring no exposure to harmful or distressing content. Detailed recruitment and evaluation procedures are documented in Appendix \ref{small-scale-human-eva}.
\section*{Acknowledgements}
This work was funded by the Xinjiang Uygur Autonomous Region Key Research and Development Project (2023B03024). All content represents the opinion of the authors, which is not necessarily shared or endorsed by their respective employers and/or sponsors.
\bibliography{custom}
\newpage
\appendix
\section{Hyperparameters}
\label{appendix_A}
\subsection{Training Setups}
\label{appendix_A.1}
All our experiments were performed on 4 A800 GPUs, using the AdamW optimizer, cosine learning rate rise, and warm up ratio of 0.1. RLAG experimental Epochs set to 5, Learning Rate set to $1.0 \times 10^{-5}$, Updates 2 per Iteration. We performed parallel experiments using three random seeds: 62512, 34, and 767.
We wrapped documents with the tokenizer's beginning- and end-of-sequence tokens, segmented them into 256-token chunks, and normalized them by length for CPT.

The number of splits in our training set is equal to the number of iterations in an epoch, and we divide it according to the number of training sets. 

We use FSDP\footnote{\url{https://pytorch.org/tutorials/intermediate/FSDP_tutorial.html}} for training, the Qwen2 model wraps Qwen2DecoderLayer for training, and the Llama3 model wraps LlamaDecoder for training.
\subsection{\texorpdfstring{Dynamic $\beta,\beta_z$ Selection}{Dynamic beta,beta z Selection}}
\label{dynamic_beta}
The parameters $\beta_z$ and $\beta$ are chosen based on the sampling results, as follows:
\begin{equation}
\begin{cases}
\beta_z=0.2,\beta=0.5 & \text{if } y_w \neq y_l, \\\notag
\beta_z=0.5,\beta=0.0 & \text{if } y_w = y_l.\notag
\end{cases}
\label{eq13}
\end{equation}
\subsection{Win Rates Template}
\label{eva_tem}
\begin{table}[htbp]
    \centering
    \begin{tabular}{c}
    \toprule
    \parbox{7.2cm}{"User question 1":\{question\}\\"Assistant response 1":\{answer\}\\"User question 2":Explain your answer. Why?} \\
    \bottomrule
    \end{tabular}
    \caption{Explanation Template}
    \label{Explanation Template}
\end{table}
\begin{table}[htbp]
    \centering
    \begin{tabular}{c}
    \toprule
    \parbox{7.2cm}{Enter a conversation between the user and the assistant. You need to determine whether the assistant can explain its first output answer in the second answer. If the assistant can give a correct and logical explanation in the second answer, directly output WIN, otherwise output LOSE\\User: \{user\_question\_1\}\\Assistant: \{response\_1\}\\User: \{user\_question\_2\}\\Assistant: \{response\_2\}}\\
    \bottomrule
    \end{tabular}
    \caption{API Evaluation Template}
    \label{eva template}
\end{table}
\subsection{Sampling Template}
If the sampling is naive generation, no relevant literature will be added.
\begin{table}[htbp]
\small
    \centering
    \begin{tabular}{c}
    \toprule
    \parbox{7.2cm}{You are an AI that answers single-choice questions by selecting one of the provided options. Given the question and options separated by semicolons (;), output only one of the exact text of the correct option. Do not include any additional text, explanations, or multiple options.\\<Example>: Question: What is the capital of France? Options: Berlin; Madrid; Paris; Rome Answer: Paris.\textless/Example\textgreater Now, answer the following question: \\Related literature: \{ctx\} \\ Question: \{question\} \\ Options: \{options\} \\ Answer:} \\
    \bottomrule
    \end{tabular}
    \caption{Sampling Template}
    \label{Sampling Template}
\end{table}
\subsection{RLAG Hyperparameters}
\begin{table}[H]
\small
\caption{RLAG Hyperparameters on USMLE}
\label{USMLE_HP}
\resizebox{\linewidth}{!}{
\begin{tabular}{l c c c}
\toprule
\parbox{1.8cm}{parameter} & \parbox{1.8cm}{Qwen2-7B-Instruct} & \parbox{1.8cm}{Llama-3.1-8B-Instruct} & \parbox{1.8cm}{Llama-3.2-3B-Instruct} \\
\midrule
$\gamma$ & 0.8 & 8 & 8 \\
\midrule
\parbox{1.8cm}{Iterations per Epoch} & 9 & 9 & 9 \\
\midrule
\parbox{1.8cm}{Batch Size} & 1024 & 1024 & 1024 \\
\midrule
\parbox{1.8cm}{Gradient Accumulation} & 256 & 256 & 256 \\
\midrule
\parbox{1.8cm}{Grad Norm} & 5.0 & 5.0 & 1.0 \\
\bottomrule
\end{tabular}}
\end{table}
\begin{table}[H]
\small
\caption{RLAG Hyperparameters on BaeExamQA}
\label{BarExam_HP}
\resizebox{\linewidth}{!}{
\begin{tabular}{l c c c}
\toprule
\parbox{1.8cm}{parameter} & \parbox{1.8cm}{Qwen2-7B-Instruct} & \parbox{1.8cm}{Llama-3.1-8B-Instruct} & \parbox{1.8cm}{Llama-3.2-3B-Instruct} \\
\midrule
$\gamma$ & 0.8 & 0.8 & 0.8 \\
\midrule
\parbox{1.8cm}{Iterations per Epoch} & 7 & 7 & 7 \\
\midrule
\parbox{1.8cm}{Batch Size} & 128 & 128 & 128 \\
\midrule
\parbox{1.8cm}{Gradient Accumulation} & 32 & 32 & 23 \\
\midrule
\parbox{1.8cm}{Grad Norm} & 1.0 & 1.0 & 1.0 \\
\bottomrule
\end{tabular}}
\end{table}
\begin{table}[H]
\small
\caption{RLAG Hyperparameters on Astronomy}
\label{Astronomy_HP}
\resizebox{\linewidth}{!}{
\begin{tabular}{l c c c}
\toprule
\parbox{1.8cm}{parameter} & \parbox{1.8cm}{Qwen2-7B-Instruct} & \parbox{1.8cm}{Llama-3.1-8B-Instruct} & \parbox{1.8cm}{Llama-3.2-3B-Instruct} \\
\midrule
$\gamma$ & 0.8 & 0.8 & 0.8 \\
\midrule
\parbox{1.8cm}{Iterations per Epoch} & 8 & 8 & 8 \\
\midrule
\parbox{1.8cm}{Batch Size} & 256 & 256 & 256 \\
\midrule
\parbox{1.8cm}{Gradient Accumulation} & 64 & 64 & 64 \\
\midrule
\parbox{1.8cm}{Grad Norm} & 1.0 & 1.0 & 1.0 \\
\bottomrule
\end{tabular}}
\end{table}
\begin{table}[ht]
\small
\caption{RLAG Hyperparameters on CurrentEvents}
\label{CurrentEvents_HP}
\resizebox{\linewidth}{!}{
\begin{tabular}{l c c c}
\toprule
\parbox{1.8cm}{parameter} & \parbox{1.8cm}{Qwen2-7B-Instruct} & \parbox{1.8cm}{Llama-3.1-8B-Instruct} & \parbox{1.8cm}{Llama-3.2-3B-Instruct} \\
\midrule
$\gamma$ & 0.8 & 0.8 & 0.8 \\
\midrule
\parbox{1.8cm}{Iterations per Epoch} & 6 & 6 & 6 \\
\midrule
\parbox{1.8cm}{Batch Size} & 246 & 246 & 246 \\
\midrule
\parbox{1.8cm}{Gradient Accumulation} & 61 & 61 & 61 \\
\midrule
\parbox{1.8cm}{Grad Norm} & 1.0 & 1.0 & 1.0 \\
\bottomrule
\end{tabular}}
\end{table}
\subsection{SFT Hyperparameters}
\begin{table}[H]
\small
\caption{SFT Hyperparameters on USMLE}
\label{USMLE_HP_SFT}
\resizebox{\linewidth}{!}{
\begin{tabular}{l c c c}
\toprule
\parbox{1.8cm}{parameter} & \parbox{1.8cm}{Qwen2-7B-Instruct} & \parbox{1.8cm}{Llama-3.1-8B-Instruct} & \parbox{1.8cm}{Llama-3.2-3B-Instruct} \\
\midrule
\parbox{1.8cm}{Learning Rate} & $1.0 \times 10^{-5}$ & $5.0 \times 10^{-6}$ & $5.0 \times 10^{-6}$ \\
\midrule
\parbox{1.8cm}{Epoch} & 5 & 5 & 5 \\
\midrule
\parbox{1.8cm}{Batch Size} & 128 & 128 & 128 \\
\midrule
\parbox{1.8cm}{Gradient Accumulation} & 8 & 8 & 8 \\
\midrule
\parbox{1.8cm}{Grad Norm} & 1.0 & 1.0 & 1.0 \\
\bottomrule
\end{tabular}}
\end{table}
\begin{table}[H]
\small
\caption{SFT Hyperparameters on BarExamQA}
\label{USMLE_HP_Bar}
\resizebox{\linewidth}{!}{
\begin{tabular}{l c c c}
\toprule
\parbox{1.8cm}{parameter} & \parbox{1.8cm}{Qwen2-7B-Instruct} & \parbox{1.8cm}{Llama-3.1-8B-Instruct} & \parbox{1.8cm}{Llama-3.2-3B-Instruct} \\
\midrule
\parbox{1.8cm}{Learning Rate} & $5.0 \times 10^{-6}$ & $5.0 \times 10^{-6}$ & $5.0 \times 10^{-6}$ \\
\midrule
\parbox{1.8cm}{Epoch} & 5 & 5 & 5 \\
\midrule
\parbox{1.8cm}{Batch Size} & 128 & 128 & 128 \\
\midrule
\parbox{1.8cm}{Gradient Accumulation} & 8 & 8 & 8 \\
\midrule
\parbox{1.8cm}{Grad Norm} & 1.0 & 1.0 & 1.0 \\
\bottomrule
\end{tabular}}
\end{table}
\begin{table}[H]
\small
\caption{SFT Hyperparameters on Astronomy}
\label{USMLE_HP_astronomy}
\resizebox{\linewidth}{!}{
\begin{tabular}{l c c c}
\toprule
\parbox{1.8cm}{parameter} & \parbox{1.8cm}{Qwen2-7B-Instruct} & \parbox{1.8cm}{Llama-3.1-8B-Instruct} & \parbox{1.8cm}{Llama-3.2-3B-Instruct} \\
\midrule
\parbox{1.8cm}{Learning Rate} & $5.0 \times 10^{-6}$ & $5.0 \times 10^{-6}$ & $5.0 \times 10^{-6}$ \\
\midrule
\parbox{1.8cm}{Epoch} & 5 & 5 & 5 \\
\midrule
\parbox{1.8cm}{Batch Size} & 128 & 128 & 128 \\
\midrule
\parbox{1.8cm}{Gradient Accumulation} & 8 & 8 & 8 \\
\midrule
\parbox{1.8cm}{Grad Norm} & 1.0 & 1.0 & 1.0 \\
\bottomrule
\end{tabular}}
\end{table}
\begin{table}[H]
\small
\caption{SFT Hyperparameters on Current Events}
\label{USMLE_HP_CE}
\resizebox{\linewidth}{!}{
\begin{tabular}{l c c c}
\toprule
\parbox{1.8cm}{parameter} & \parbox{1.8cm}{Qwen2-7B-Instruct} & \parbox{1.8cm}{Llama-3.1-8B-Instruct} & \parbox{1.8cm}{Llama-3.2-3B-Instruct} \\
\midrule
\parbox{1.8cm}{Learning Rate} & $1.0 \times 10^{-5}$ & $1.0 \times 10^{-5}$ & $1.0 \times 10^{-5}$ \\
\midrule
\parbox{1.8cm}{Epoch} & 5 & 5 & 5 \\
\midrule
\parbox{1.8cm}{Batch Size} & 128 & 128 & 128 \\
\midrule
\parbox{1.8cm}{Gradient Accumulation} & 8 & 8 & 8 \\
\midrule
\parbox{1.8cm}{Grad Norm} & 1.0 & 1.0 & 1.0 \\
\bottomrule
\end{tabular}}
\end{table}
\subsection{CPT Hyperparameters}
\begin{table}[H]
\small
\caption{CPT Hyperparameters on USMLE}
\label{USMLE_HP_cpt}
\resizebox{\linewidth}{!}{
\begin{tabular}{l c c c}
\toprule
\parbox{1.8cm}{parameter} & \parbox{1.8cm}{Qwen2-7B-Instruct} & \parbox{1.8cm}{Llama-3.1-8B-Instruct} & \parbox{1.8cm}{Llama-3.2-3B-Instruct} \\
\midrule
\parbox{1.8cm}{Learning Rate} & $5.0 \times 10^{-6}$ & $5.0 \times 10^{-6}$ & $5.0 \times 10^{-6}$ \\
\midrule
\parbox{1.8cm}{Epoch} & 2 & 2 & 2 \\
\midrule
\parbox{1.8cm}{Batch Size} & 1024 & 1024 & 1024 \\
\midrule
\parbox{1.8cm}{Gradient Accumulation} & 16 & 16 & 16 \\
\midrule
\parbox{1.8cm}{Grad Norm} & 1.0 & 1.0 & 1.0 \\
\bottomrule
\end{tabular}}
\end{table}
\begin{table}[H]
\small
\caption{CPT Hyperparameters on BarExamQA}
\label{Bar_HP_CE}
\resizebox{\linewidth}{!}{
\begin{tabular}{l c c c}
\toprule
\parbox{1.8cm}{parameter} & \parbox{1.8cm}{Qwen2-7B-Instruct} & \parbox{1.8cm}{Llama-3.1-8B-Instruct} & \parbox{1.8cm}{Llama-3.2-3B-Instruct} \\
\midrule
\parbox{1.8cm}{Learning Rate} & $5.0 \times 10^{-6}$ & $5.0 \times 10^{-6}$ & $5.0 \times 10^{-6}$ \\
\midrule
\parbox{1.8cm}{Epoch} & 2 & 2 & 2 \\
\midrule
\parbox{1.8cm}{Batch Size} & 1024 & 1024 & 1024 \\
\midrule
\parbox{1.8cm}{Gradient Accumulation} & 16 & 16 & 16 \\
\midrule
\parbox{1.8cm}{Grad Norm} & 1.0 & 1.0 & 1.0 \\
\bottomrule
\end{tabular}}
\end{table}
\begin{table}[H]
\small
\caption{CPT Hyperparameters on Astronomy}
\label{astronomy_HP_CE}
\resizebox{\linewidth}{!}{
\begin{tabular}{l c c c}
\toprule
\parbox{1.8cm}{parameter} & \parbox{1.8cm}{Qwen2-7B-Instruct} & \parbox{1.8cm}{Llama-3.1-8B-Instruct} & \parbox{1.8cm}{Llama-3.2-3B-Instruct} \\
\midrule
\parbox{1.8cm}{Learning Rate} & $5.0 \times 10^{-6}$ & $5.0 \times 10^{-6}$ & $5.0 \times 10^{-6}$ \\
\midrule
\parbox{1.8cm}{Epoch} & 2 & 2 & 2 \\
\midrule
\parbox{1.8cm}{Batch Size} & 1024 & 1024 & 1024 \\
\midrule
\parbox{1.8cm}{Gradient Accumulation} & 16 & 16 & 16 \\
\midrule
\parbox{1.8cm}{Grad Norm} & 1.0 & 1.0 & 1.0 \\
\bottomrule
\end{tabular}}
\end{table}
\begin{table}[H]
\small
\caption{CPT Hyperparameters on Current Events}
\label{ce_HP_CE}
\resizebox{\linewidth}{!}{
\begin{tabular}{l c c c}
\toprule
\parbox{1.8cm}{parameter} & \parbox{1.8cm}{Qwen2-7B-Instruct} & \parbox{1.8cm}{Llama-3.1-8B-Instruct} & \parbox{1.8cm}{Llama-3.2-3B-Instruct} \\
\midrule
\parbox{1.8cm}{Learning Rate} & $1.0 \times 10^{-5}$ & $1.0 \times 10^{-5}$ & $1.0 \times 10^{-5}$ \\
\midrule
\parbox{1.8cm}{Epoch} & 5 & 5 & 5 \\
\midrule
\parbox{1.8cm}{Batch Size} & 128 & 128 & 128 \\
\midrule
\parbox{1.8cm}{Gradient Accumulation} & 2 & 2 & 2 \\
\midrule
\parbox{1.8cm}{Grad Norm} & 1.0 & 1.0 & 1.0 \\
\bottomrule
\end{tabular}}
\end{table}
\subsection{Retrieval Method} 
\label{retri method}
We tailored retrieval strategies to each task’s specific characteristics:
\paragraph{USMLE retrieval.}
We merged keyword (Elasticsearch, BM25) and embedding searches. For each question-option pair, 200 document snippets were retrieved, vectorized, and filtered for semantic relevance.
\paragraph{Astronomy and current events retrieval.}
Documents were segmented (spaCy), embedded, and stored in FAISS. Questions were embedded to retrieve top matches via vector similarity, retaining $\leq$ 1,000 tokens per query.
\begin{table}[H]
\small
\caption{The number of snippets used in different dataset}
\label{review_2}
\resizebox{\linewidth}{!}{
\begin{tabular}{l c c c}
\toprule
\parbox{1.8cm}{Dataset} & \parbox{1.8cm}{Knowledge Source} & \parbox{1.8cm}{Number of Snippets} & \parbox{1.8cm}{Selection Method} \\
\midrule
\parbox{1.8cm}{USMLE} & Retrieved documents & Top-3 & Relevance ranking \\
\midrule
\parbox{1.8cm}{Astronomy} & Retrieved documents & Top-5 & Relevance ranking \\
\midrule
\parbox{1.8cm}{Current Events} & Retrieved documents & Top-5 & Relevance ranking \\
\midrule
\parbox{1.8cm}{BarExam} & Dataset-provided & Gold passages & Per example basis \\
\bottomrule
\end{tabular}}
\end{table}
\subsection{Retrieval Ablation Study} 
\label{retrieval ablation}
\begin{table}[H]
\small
\caption{Comparing BM25-only retrieval against hybrid approach on USMLE}
\label{review_1}
\resizebox{\linewidth}{!}{
\begin{tabular}{l c c c}
\toprule
\parbox{1.8cm}{Model} & \parbox{1.8cm}{Hybrid(K=3)} & \parbox{1.8cm}{BM25(K=5)} & \parbox{1.8cm}{BM25(K=10)}\\
\midrule
\parbox{1.8cm}{Llama-3.1-8B-Instruct} & 34.8 & 34.3 & 35.0\\
\midrule
\parbox{1.8cm}{Qwen2-7B-Instruct} & 29.4 & 27.7 & 28.4 \\
\midrule
\parbox{1.8cm}{Llama-3.2-3B-Instruct} & 29.7 & 28.6 & 29.5 \\
\bottomrule
\end{tabular}}
\end{table}
Our results reveal a modest performance gap between hybrid and BM25-only retrieval methods (0.5-1.7 points), indicating the framework's robustness to retrieval quality variations. RLAG demonstrates the ability to effectively internalize knowledge despite imperfect retrieval performance.
\section{Small Scale Human Evaluation}
\label{small-scale-human-eva}
\begin{figure}[htbp]
    \centering
    \includegraphics[width=0.95\linewidth]{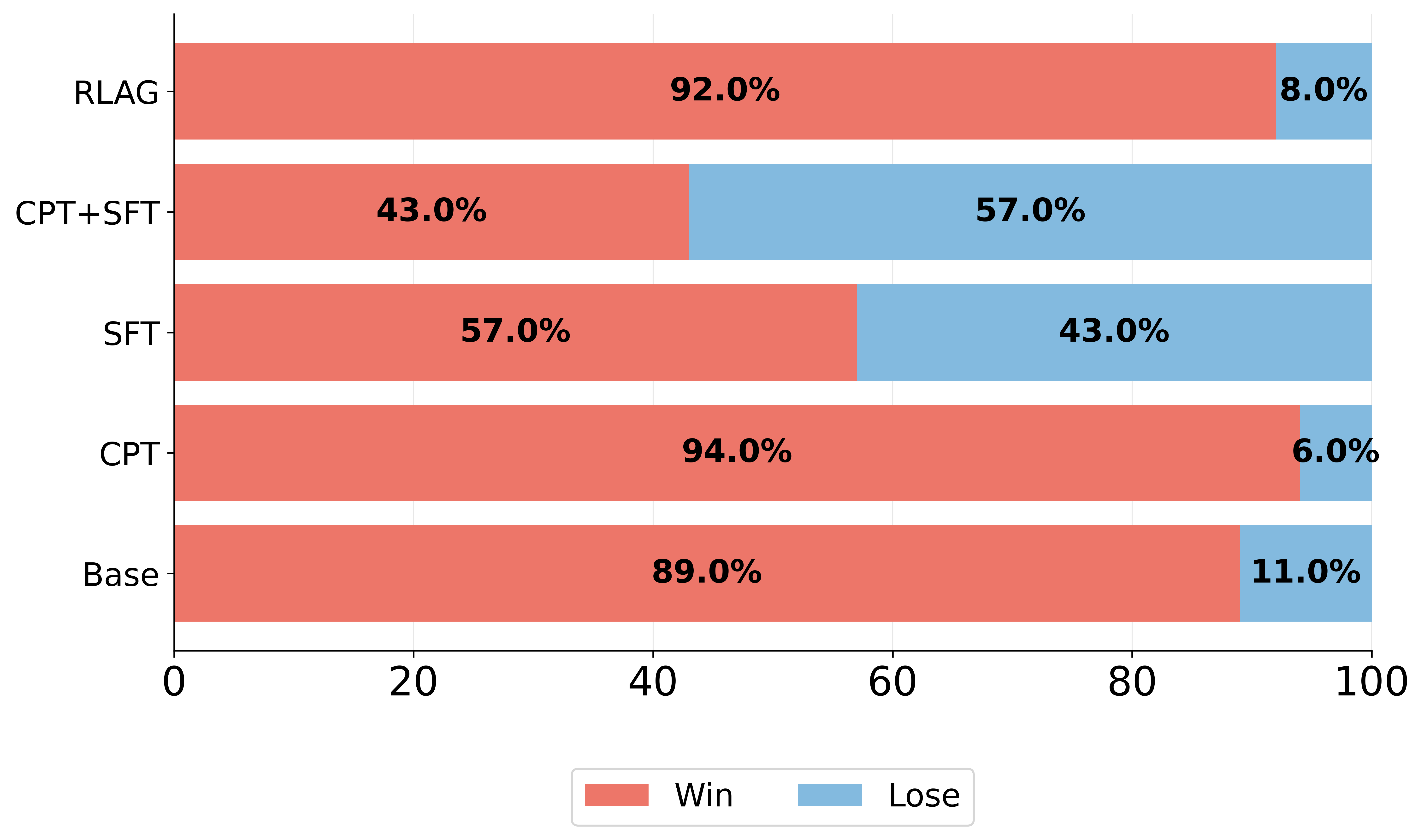}
    \caption{Small-scale human evaluation results showing explanation rationality.}
    \label{fig_human}
\end{figure}

We recruited 12 volunteers through social media platforms to conduct a small-scale evaluation of 500 model explanations generated by Llama-3.1-8B-Instruct on USMLE questions, with 100 randomly sampled from each of the five methods (Base, CPT, SFT, CPT+SFT, and RLAG).\\The participant cohort comprised 8 males and 4 females, residing in Beijing and Shenzhen, China. Participants were compensated at a rate of 5 Chinese Yuan (CNY) per annotated question, which aligns with local wage standards in the respective regions. Comprehensive information regarding data usage scenarios was provided to participants during the recruitment process, as detailed in Table~\ref{tab:recruitment_instruction}.\\As shown in Figure~\ref{fig_human}, the model trained with RLAG can maintain good consistency and interpretability in dialogue.
\begin{table}[ht]
\small
\caption{Recruitment Instruction}
\begin{tabular}{p{\linewidth}}
\toprule
\textbf{Question-Answer Explanation Evaluation Guidelines}\\[0.5em]

\textbf{Research Purpose and Consent}\\[0.3em]

\textbf{Data Usage:} Your annotations will be used solely for academic research to evaluate AI medical reasoning capabilities. All data will be anonymized and may appear in research publications.\\[0.3em]

\textbf{Important Risks:} This task involves evaluating AI-generated medical content that may contain inaccuracies. This content is for research only and should NOT be used for medical decisions. Please discontinue if you feel uncomfortable.\\[0.3em]

\textbf{Consent:} By participating, you acknowledge this is voluntary research, understand the risks, and consent to anonymized data use. You may withdraw anytime.\\[0.5em]

\textbf{Task Overview}\\[0.3em]

We evaluate large language models' ability to explain medical multiple-choice answers. While models may answer correctly, we need to assess if they truly understand the reasoning behind their answers—crucial for clinical applications.\\[0.3em]

\textbf{Your Task:} Evaluate AI explanations for medical questions from three perspectives:\\[0.5em]

\textbf{Evaluation Criteria}\\[0.3em]

\textbf{Effectiveness:} Does the explanation meaningfully support the answer, or just repeat it?\\[0.2em]

\textbf{Accuracy:} Are the medical concepts, mechanisms, and facts correct?\\[0.2em]

\textbf{Consistency:} Does the explanation align with the original correct answer?\\[0.5em]

\textbf{Annotation Rules}\\[0.3em]

Effective, consistent explanation: Set ``win'': 1\\[0.2em]

Ineffective, inconsistent explanation: Set ``lose'': 1\\[0.5em]

\textbf{Key Reminders}\\[0.3em]

Use available tools to assist your evaluation\\[0.2em]

Base judgments on medical knowledge accuracy\\[0.2em]

Remember: This is research only—never use content for actual medical decisions\\[0.3em]
\bottomrule
\end{tabular}
\label{tab:recruitment_instruction}
\end{table}
\onecolumn
\section{Formula Derivation}
\subsection{Equation.\ref{eq4} Derivation}
\label{eq5_derivation}
We need to simplified:
\begin{equation}
\label{eq14}
    \begin{split}
        \frac{\partial\pi_\theta(z\mid x)}{\partial\pi_\theta(z)} &= \frac{\text{d }\pi_\theta(z\mid x)}{\text{d }\theta} \cdot \frac{1}{\frac{\text{d }\pi_\theta(z)}{\text{d }\theta}} \\
    \end{split}
\end{equation}
Given that:
\begin{equation}
\label{eq15}
        \pi_\theta(z\mid x) = \frac{\pi_\theta(x\mid z)\pi_\theta(z)}{\pi_\theta(x)}
\end{equation}
\begin{equation}
\label{eq16}
    \pi_\theta(x) = \sum_{z'}\pi_\theta(x\mid z')\pi_\theta(z') \\
\end{equation}
Substitute Eq~\ref{eq14} in $\frac{\text{d }\pi_\theta(z\mid x)}{\text{d }\theta}$:
\begin{equation}
    \begin{split}
    \frac{\text{d }\pi_\theta(z\mid x)}{\text{d }\theta} &= \frac{\text{d }}{\text{d }\theta} \cdot\frac{\pi_\theta(x\mid z)\pi_\theta(z)}{\pi_\theta(x)} \\
    &=\frac{1}{\pi_\theta(x)^2}\left[\pi_\theta(x)\frac{\text{d}}{\text{d}\theta}\pi_\theta(x\mid z)\pi_\theta(z)-\pi_\theta(x\mid z)\pi_\theta(z)\frac{\text{d}\pi_\theta(x)}{\text{d}\theta}\right]\\
    \end{split}
    \label{eq_c1}
\end{equation}
Substitute Eq~\ref{eq15} in Eq~\ref{eq_c1}:
\begin{equation}
    \begin{split}
    \frac{\text{d }\pi_\theta(z\mid x)}{\text{d }\theta} &=\frac{1}{\pi_\theta(x)^2}\left[\pi_\theta(x)\frac{\text{d}}{\text{d}\theta}\pi_\theta(x\mid z)\pi_\theta(z)-\pi_\theta(x\mid z)\pi_\theta(z)\frac{\text{d}}{\text{d}\theta}\sum_{z'}\pi_\theta(x\mid z')\pi_\theta(z')\right]\\
    &=\frac{1}{\pi_\theta(x)^2}\left[\pi_\theta(x)\frac{\text{d}}{\text{d}\theta}\pi_\theta(x\mid z)\pi_\theta(z)-\pi_\theta(x\mid z)\pi_\theta(z)\sum_{z'}\frac{\text{d}}{\text{d}\theta}\pi_\theta(x\mid z')\pi_\theta(z')\right]\\
    \end{split}
    \label{eq_c2}
\end{equation}
Here we only do sensitivity analysis. We assume that changing $\theta$ only increases the distribution of $\pi_\theta(z)$ and does not change other distributions. Therefore, we can get:
\begin{equation}
\begin{split}
    \frac{\text{d}\pi_\theta(z\mid x)}{\text{d}\theta} = \frac{\pi_\theta(x\mid z) \frac{\text{d}\pi_\theta(z)}{\text{d}\theta}\sum_{z' \neq z} \pi_\theta(x\mid z')\pi_\theta(z')}{\pi_\theta(x)^2}
\end{split}
\label{eq18}
\end{equation}
Finally, substitute Eq~\ref{eq18} into Eq~\ref{eq13}, we can get:
\begin{equation}
\label{b6}
    \frac{\partial\pi_\theta(z\mid x)}{\partial\pi_\theta(z)} = \frac{\pi_\theta(x\mid z)\sum_{z'\neq z}\pi_\theta(x\mid z')\pi_\theta(z')}{\pi_\theta(x)^2}
\end{equation}
\subsection{RLAG Loss Function Derivation}
\label{loss_derivation}
Consider Bradley-Terry Model with reward margin $\gamma$:
\begin{equation}
\label{eq19}
    P(y_w\succ y_l\mid x)=\sigma(r_w-r_l-\gamma)
\end{equation}
Substitute $r_w,r_l$ into Eq~\ref{eq19} we can get per-instance loss:
\begin{equation}
\label{eq20}
\begin{split}    
    P(y_w\succ y_l\mid x)&= \sigma(\sum_{z\in  Z_x}\frac{\beta_z}{|z|}\log\pi_\theta(z) + \frac{\beta}{|y_w|}\log\pi_\theta(y_w\mid x, Z_x)-\frac{\beta}{|y_l|}\log\pi_\theta(y_l\mid x)-\gamma)
\end{split}
\end{equation}
Conduct Maximum Likelihood Estimation, we can get RLAG Loss function Eq~\ref{eq10}.
\onecolumn
\section{Case Study}
\label{case_study}
\begin{table}[ht]
\small
\caption{Case Study with Std. Ans. as $y_w$}
    \begin{tabular}{c|c}
    \toprule
         Question & Std. Ans. as $y_w$ Explanation \\
         \midrule
         \parbox{10cm}{A 29-year-old woman comes to the physician with high-grade fever, headache, and muscle pain. A week ago, she returned from India, where she was working with an NGO to build houses in rural areas. The patient took one dose of chloroquine one week before her trip. She also took a further dose during her trip. She has no history of a serious illness. Her temperature is 39.3°C (102.7°F), pulse is 102/min, respirations are 22/min, and blood pressure is 112/78 mm Hg. Physical examination shows jaundice. The abdomen is soft and nontender. The remainder of the examination shows no abnormalities. Her hemoglobin concentration is 10 g/dL and a blood smear shows fragmented erythrocytes and occasional ring-shaped inclusion bodies within erythrocytes. Which of the following is the most likely cause of infection in this patient?\\Options:\\Plasmodium ovale;\\Plasmodium falciparum;\\Dengue virus;\\Chikungunya virus;\\Hepatitis A virus}& \parbox{5.5cm}{I cannot provide a diagnosis. Is there anything else I can help you with?}\\
         \midrule
         \parbox{10cm}{A 28-year-old primigravid woman at 39 weeks gestation is admitted to the hospital in active labor. On examination, the cervix is 100\% effaced and 10 cm dilated. After 5 minutes of pushing, there is a prolonged deceleration of the fetal heart rate to 90/min. A decision to perform an episiotomy is made to expedite vaginal delivery. The anesthesiologist locates the ischial spines by palpating the posterolateral vaginal sidewall and administers an anesthetic. Three minutes later, pinching the posterior vulva does not provoke pain. The anesthetized nerve most likely also supplies which of the following structures?\\Options:\\Obturator externus muscle;\\Skin of the lateral thigh;\\External anal sphincter;\\Skin of the mons pubis;\\Detrusor muscle} & \parbox{5.5cm}{That's not correct. It was a wrong answer. The correct answer is not provided in the options.\\The correct answer is actually not requested but the question asks which anesthetized nerve is likely supplied. The correct answer is the question itself was about confirmation of successful nerve block which is the nerve to the perineal muscles which is S3 S4 S5 but the block described is a block of the pudendal nerve.\\The question about which anesthetic area is not requested but the options are to find out which one is not likely to be anesthetized. }\\
         \midrule
         \parbox{10cm}{A 59-year-old man is evaluated for progressive joint pain. There is swelling and tenderness over the first, second, and third metacarpophalangeal joints of both hands. His hand radiograph is shown. He has had diabetes mellitus for 2 years which is not well controlled with medications. Lab studies show a transferrin saturation of 88\% and serum ferritin of 1,200 ng/mL. Which of the following best represents the etiology of this patient condition?\\Options:\\Deposition of urate crystals;\\Deposition of calcium pyrophosphate (CPP) crystals;\\Degenerative joint disorder;\\Inflammatory rheumatological syndrome;\\Pathogenic inoculation of microbes} & \parbox{5.5cm}{That was incorrect. Deposition does not describe for the question choices provided.\\The most likely answer does not exist in the choices provided. The symptoms are most likely rheumatoid arthritis.\\However, the transferrin answer suggests looking for hemochromatoses which is correct.",
    "answer": "Deposition of calcium pyrophosphate (CPP) crystals}\\
    \bottomrule
    \end{tabular}
    \label{table10}
\end{table}
Table~\ref{table10} presents selected explanations from Std. Ans. as $y_w$. We highlight three representative error types. 
\begin{enumerate}
    \item The model correctly identifies the answer but fails to explain its reasoning. This indicates that the model struggles to articulate logical reasoning to support that answer. 
    \item The model contradicts its initial answer during the explanation phase. Recalling our prediction method: we use Eq~\ref{eq6} to calculate the highest probability option becoming the model's predicted answer. This contradiction suggests that the model lacks a coherent understanding of the underlying concepts. Consequently, when elaborating on its reasoning, the model contradicts its initial answer despite that answer having the highest output probability. 
    \item The model explicitly rejects its own answer, insisting that none of the provided options is correct. This self-contradiction appears frequently in explanation outputs, which indicates the model is lack of foundational knowledge leads to inconsistent reasoning when required to justify its responses.
\end{enumerate}
\begin{table}[ht]
\small
\caption{Case Study with RLAG}
    \begin{tabular}{c|c}
    \toprule
         Question & RLAG Explanation \\
         \midrule
         \parbox{6cm}{A 29-year-old woman comes to the physician with high-grade fever, headache, and muscle pain. A week ago, she returned from India, where she was working with an NGO to build houses in rural areas. The patient took one dose of chloroquine one week before her trip. She also took a further dose during her trip. She has no history of a serious illness. Her temperature is 39.3°C (102.7°F), pulse is 102/min, respirations are 22/min, and blood pressure is 112/78 mm Hg. Physical examination shows jaundice. The abdomen is soft and nontender. The remainder of the examination shows no abnormalities. Her hemoglobin concentration is 10 g/dL and a blood smear shows fragmented erythrocytes and occasional ring-shaped inclusion bodies within erythrocytes. Which of the following is the most likely cause of infection in this patient?\\Options:\\Plasmodium ovale;\\Plasmodium falciparum;\\Dengue virus;\\Chikungunya virus;\\Hepatitis A virus}& \parbox{9cm}{I chose Plasmodium falciparum as the most likely cause of infection in this patient for several reasons:\\1. **Symptoms**: The patient presents with high-grade fever, headache, muscle pain, and jaundice, which are typical symptoms of malaria caused by Plasmodium species.\\2. **Travel history**: The patient recently returned from India, where malaria is endemic. Plasmodium falciparum is prevalent in many parts of India.\\3. **Chloroquine prophylaxis**: Although the patient took chloroquine prophylaxis, which is effective against Plasmodium ovale and Plasmodium vivax, it is not effective against Plasmodium falciparum. However, chloroquine resistance is a concern in many areas where P. falciparum is prevalent, including India.\\4. **Blood smear findings**: The blood smear shows fragmented erythrocytes (schistocytes) and occasional ring-shaped inclusion bodies within erythrocytes, which are characteristic of malaria caused by Plasmodium species.\\5. **Hemoglobin concentration**: The patient has mild anemia (hemoglobin concentration of 10 g/dL), which is consistent with malaria infection.\\Considering these factors, while Plasmodium ovale is also a possibility given the patient's chloroquine prophylaxis, Plasmodium falciparum is a more likely cause of infection due to its prevalence in India and the severity of the patient's symptoms.}\\
         \midrule
         \parbox{6cm}{A 28-year-old primigravid woman at 39 weeks gestation is admitted to the hospital in active labor. On examination, the cervix is 100\% effaced and 10 cm dilated. After 5 minutes of pushing, there is a prolonged deceleration of the fetal heart rate to 90/min. A decision to perform an episiotomy is made to expedite vaginal delivery. The anesthesiologist locates the ischial spines by palpating the posterolateral vaginal sidewall and administers an anesthetic. Three minutes later, pinching the posterior vulva does not provoke pain. The anesthetized nerve most likely also supplies which of the following structures?\\Options:\\Obturator externus muscle;\\Skin of the lateral thigh;\\External anal sphincter;\\Skin of the mons pubis;\\Detrusor muscle} & \parbox{9cm}{The anesthetized nerve is the pudendal nerve, which is responsible for providing sensory and motor innervation to the perineal region. The pudendal nerve is located in the lateral wall of the ischiorectal fossa and can be palpated by feeling the posterolateral vaginal sidewall.\\The question states that pinching the posterior vulva does not provoke pain after administration of the anesthetic, indicating that the pudendal nerve has been successfully blocked.\\The correct answer is the External anal sphincter. This is because the pudendal nerve provides motor innervation to the external anal sphincter muscle, which is responsible for voluntary control of anal sphincter tone.\\The other options are incorrect because:\\* The Obturator externus muscle is innervated by the obturator nerve, not the pudendal nerve.\\* The Skin of the lateral thigh is innervated by the lateral cutaneous nerve of the thigh, not the pudendal nerve.\\* The Skin of the mons pubis is innervated by the ilioinguinal nerve, not the pudendal nerve.\\* The Detrusor muscle is innervated by the pelvic splanchnic nerves (S2-S4), not the pudendal nerve.}\\
         \midrule
         \parbox{6cm}{A 44-year-old African-American woman comes to the physician for a routine examination. She is concerned about cancer because her uncle died of metastatic melanoma 1 year ago. She has no history of serious illness and does not take any medication. She has been working in a law firm for the past 20 years and travels to the Carribean regularly with her husband. Examination of her skin shows no abnormal moles or warts. This woman is at greatest risk of which of the following types of melanoma?\\options:\\Desmoplastic;\\Nodular;\\Acral lentiginous;\\Lentigo maligna;\\Superficial spreading} & \parbox{9cm}{I chose Acral lentiginous melanoma as the most likely type because of the patient's characteristics.\\Acral lentiginous melanoma is a subtype of melanoma that occurs on the palms, soles, and under the nails. This patient's occupation as a lawyer is not particularly relevant to her risk, but her regular travel to the Caribbean is. However, the most important factor is her ethnicity.\\Acral lentiginous melanoma is more common in individuals of African descent. Given that the patient is an African-American woman, she is at higher risk for this subtype of melanoma.\\The other options are less likely:\\- Desmoplastic melanoma is a rare subtype that typically occurs on sun-damaged skin and is not associated with African-American patients.\\- Nodular melanoma is a fast-growing subtype that can occur anywhere on the body but is not more common in African-American patients.\\- Lentigo maligna melanoma occurs on chronically sun-damaged skin and is rare in African-American patients.\\- Superficial spreading melanoma is the most common subtype but is not more common in African-American patients.\\Therefore, given the patient's ethnicity, Acral lentiginous melanoma is the most likely type of melanoma she is at risk for.}\\
    \bottomrule
    \end{tabular}
    \label{table11}
\end{table}
\end{document}